\documentclass[11pt,a4paper]{article}
\usepackage[utf8]{inputenc}
\usepackage[T1]{fontenc}
\usepackage[english]{babel}
\usepackage{amsmath,amssymb}
\usepackage{graphicx}
\usepackage{multirow}
\usepackage{indentfirst}
\usepackage{hyperref}
\usepackage{pifont}
\usepackage{appendix}

\usepackage[table]{xcolor}
\usepackage{listings}
\usepackage[margin=1in]{geometry}
\usepackage{caption}
\usepackage{subcaption}
\usepackage{fontawesome5}
\usepackage{longtable}
\usepackage{enumitem}
\usepackage{mdframed}
\usepackage{colortbl}
\usepackage{booktabs}
\usepackage{float}
\usepackage{tabularx}
\usepackage{xltabular}
\usepackage{siunitx}
\definecolor{frontierBg}{HTML}{E8F5EC}
\definecolor{frontierLabel}{HTML}{1B7A3D}
\definecolor{openBg}{HTML}{E8E8F4}
\definecolor{openLabel}{HTML}{3D405B}
\definecolor{ptBg}{HTML}{FFF0DE}
\definecolor{ptLabel}{HTML}{9E5A08}
\definecolor{bestcell}{HTML}{D4EDDA}
\definecolor{gray60}{HTML}{999999}
\definecolor{perf1}{HTML}{F5B7B1}   
\definecolor{perf2}{HTML}{F5CBA7}   
\definecolor{perf3}{HTML}{FAE5A0}   
\definecolor{perf4}{HTML}{A9DFBF}   
\definecolor{perf5}{HTML}{7DCEA0}   
\definecolor{perf6}{HTML}{52BE80}   
\definecolor{perfTop}{HTML}{2ECC71} 

\definecolor{coh1}{HTML}{FAE5A0}    
\definecolor{coh2}{HTML}{A9DFBF}    
\definecolor{coh3}{HTML}{7DCEA0}    
\definecolor{coh4}{HTML}{52BE80}   

\usepackage{tcolorbox}
\usepackage{authblk}

\usepackage{threeparttable}
\usepackage[numbers]{natbib}

\newcommand{\cmark}{\faCheck}
\newcommand{\xmark}{\faTimes}

\definecolor{codegreen}{rgb}{0,0.6,0}
\definecolor{codegray}{rgb}{0.5,0.5,0.5}
\definecolor{codepurple}{rgb}{0.58,0,0.82}
\definecolor{backcolour}{rgb}{0.95,0.95,0.92}

\lstdefinestyle{mystyle}{
    backgroundcolor=\color{backcolour},
    commentstyle=\color{codegreen},
    keywordstyle=\color{magenta},
    numberstyle=\tiny\color{codegray},
    stringstyle=\color{codepurple},
    basicstyle=\ttfamily\footnotesize,
    breakatwhitespace=false,
    breaklines=true,
    captionpos=b,
    keepspaces=true,
    numbers=left,
    numbersep=5pt,
    showspaces=false,
    showstringspaces=false,
    showtabs=false,
    tabsize=2
}
\title{\textbf{CAPITU: A Benchmark for Evaluating Instruction-Following in Brazilian Portuguese with Literary Context}}

\author[1]{Giovana Kerche Bonás}
\author[1]{Roseval Malaquias Junior}
\author[2]{Marcos Piau}
\author[1]{Thiago Laitz}
\author[1]{Thales Sales Almeida}
\author[1]{Hugo Abonizio}
\author[2]{Celio Larcher}
\author[1]{Ramon Pires}
\author[1]{Rodrigo Nogueira}
\affil[1]{Maritaca AI}
\affil[2]{Jusbrasil}

\date{March 2026}

\begin{document}

\maketitle

\begin{abstract}
	We introduce CAPITU, a benchmark for evaluating instruction-following capabilities of Large Language Models (LLMs) in Brazilian Portuguese. Unlike existing benchmarks that focus on English or use generic prompts, CAPITU contextualizes all tasks within eight canonical works of Brazilian literature, combining verifiable instruction constraints with culturally-grounded content. The benchmark comprises 59 instruction types organized into seven categories, all designed to be automatically verifiable without requiring LLM judges or human evaluation. Instruction types include Portuguese-specific linguistic constraints (word termination patterns like -ando/-endo/-indo, -inho/-inha, -mente) and structural requirements. We evaluate 18 state-of-the-art models across single-turn and multi-turn settings. Our results show that frontier reasoning models achieve strong performance (GPT-5.2 with reasoning: 98.5\% strict accuracy), while Portuguese-specialized models offer competitive cost-efficiency (Sabiazinho-4: 87.0\% at \$0.13 vs Claude-Haiku-4.5: 73.5\% at \$1.12). Multi-turn evaluation reveals significant variation in constraint persistence, with conversation-level accuracy ranging from 60\% to 96\% across models. We identify specific challenges in morphological constraints, exact counting, and constraint persistence degradation across turns. We release the complete benchmark, evaluation code, and baseline results to facilitate research on instruction-following in Portuguese.
\end{abstract}

\section{Introduction}
\label{sec:introduction}

Large Language Models (LLMs) exhibit versatility across a broad spectrum of tasks, from creative writing to complex problem-solving. However, their practical deployment hinges on their ability to follow precise instructions without compromising output quality. Evaluating this capability through explicit constraints offers a quantifiable measure of model controllability. The capacity to satisfy rigid structural requirements validates whether a model can preserve coherence and utility under strict boundary conditions, thereby indicating greater robustness~\citep{ouyang2022training}.

The importance of instruction-following evaluation has driven the development of specialized benchmarks. IFEval~\citep{zhou2023instruction} pioneered a paradigm of verifiable instructions with 25 constraint types that can be automatically checked without subjective judgment, establishing a foundation for systematic evaluation in English. Subsequent work has expanded this paradigm: IFBench~\citep{pyatkin2025ifbench} introduced 58 new instruction types to test generalization to novel constraints, FollowBench~\citep{jiang2024followbench} organized complexity into hierarchical levels, and Multi-IF~\citep{he2024multiif} extended this approach to eight languages, including Portuguese, through the translation of English prompts.

Despite these advances, instruction-following evaluation for Portuguese remains comparatively underexplored. While Multi-IF~\citep{he2024multiif} provides a baseline for multilingual coverage, evaluation sets built primarily via direct translation may not fully reflect language-specific phenomena or culturally grounded contexts. Portuguese exhibits rich morphological features and discourse patterns that differ fundamentally from English, necessitating a language-specific instruction design rather than adaptation.

The limitations of translation-based evaluation become particularly apparent for Portuguese. First, linguistic specificity: Portuguese exhibits morphological and syntactic features that cannot be adequately assessed through translated English instructions—its productive diminutive system (-inho/-zinho), gerund constructions, and adverb formation (-mente) require native instruction design \citep{goncalves2022morphology}. Second, discourse conventions: translated instructions reflect English writing patterns and rhetorical norms, while Portuguese has distinct conventions for formality, argumentation, and connective usage. Third, evaluation authenticity: translated instructions may introduce artifacts or miss language-specific constraints central to Portuguese writing.

We address these challenges by introducing \textbf{CAPITU} (\textit{Contextual Assessment of Portuguese Instruction-following Through Literary Understanding}), an instruction-following benchmark designed natively for Brazilian Portuguese. Rather than translating existing English benchmarks, CAPITU contextualizes its prompts within eight canonical works of Brazilian literature spanning Romanticism, Realism, Naturalism, Regionalism and Modernism. This literary framing provides natural, coherent scenarios for generation tasks, while evaluation focuses on verifiable constraint satisfaction. CAPITU uses 59 instruction types, including Portuguese-specific constraints targeting morphological phenomena such as diminutive suffixes and gerund forms that lack direct English equivalents. CAPITU is available at \href{https://github.com/maritaca-ai/capitu}{https://github.com/maritaca-ai/capitu}.

A key design principle of CAPITU is automatic verification: all 59 instruction types are automatically verifiable using string operations, regular expressions, and lexicon matching. This eliminates the need for LLM judges or human annotators to evaluate instruction compliance, ensuring reproducible and objective evaluation. While LLM-as-judge approaches have shown strong correlation with human evaluation~\citep{zheng2023judging, liu2023geval}, they introduce non-determinism and cost overhead. 

CAPITU's literary grounding serves multiple purposes. Brazilian and Portuguese literary classics provide naturally occurring text with varied narrative techniques, regional varieties, and historical registers that contextualize instruction-following within linguistically rich material.

This paper is organized as follows: Section~\ref{sec:related} reviews related work in instruction-following evaluation, multilingual benchmarks, and literary Natural Language Processing (NLP). Section~\ref{sec:methodology} describes CAPITU's design principles, instruction taxonomy, and evaluation framework.  Section~\ref{sec:statistics} provides benchmark statistics. Section~\ref{sec:experiments} presents our experimental setup and baseline results across multiple models. Section~\ref{sec:results} analyzes performance patterns and failure modes. Section~\ref{sec:discussion} discusses implications, limitations, and future directions. We conclude in Section~\ref{sec:conclusion}.

\subsection{Contributions}

\begin{figure}[t]
\centering
\includegraphics[width=\textwidth]{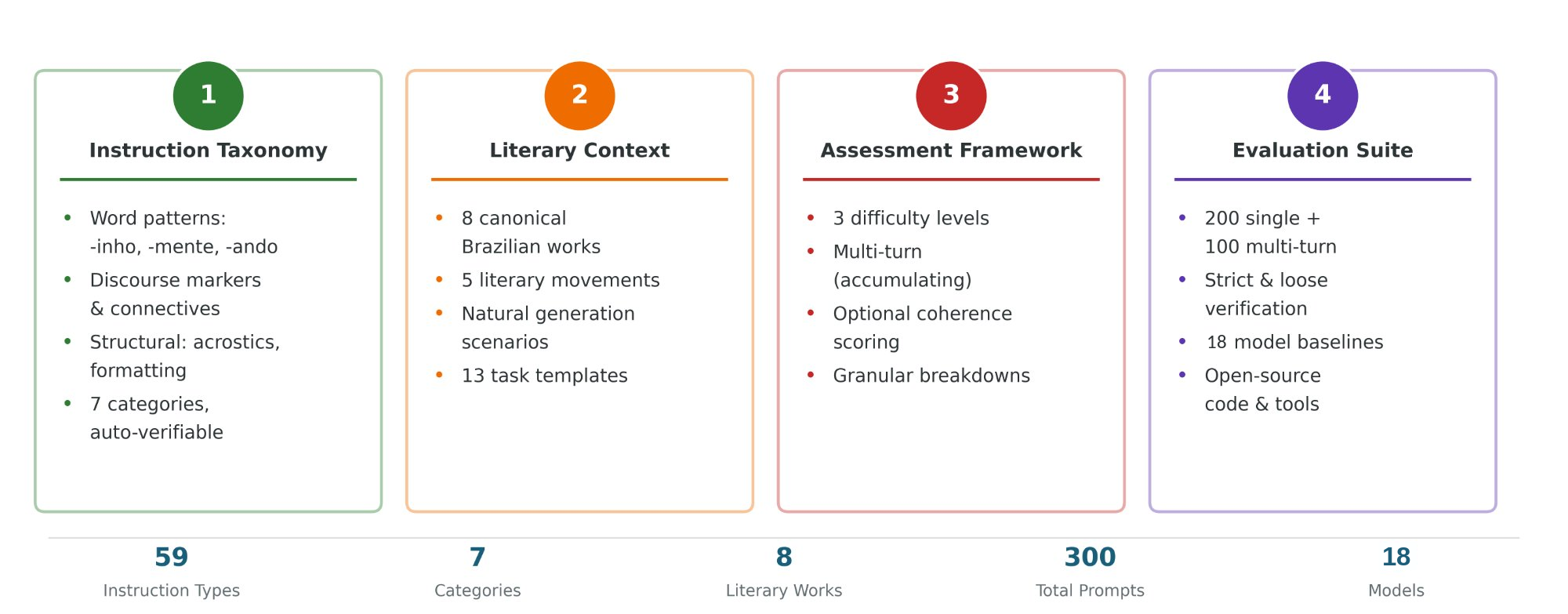}
\caption{Overview of CAPITU's main contributions.}
\label{fig:contributions}
\end{figure}

Figure~\ref{fig:contributions} summarizes our four main contributions: 
(1) a Portuguese-specific instruction taxonomy with 59 automatically 
verifiable types across 7 categories, including morphological patterns 
unique to Portuguese; (2) culturally-contextualized prompts grounded in 
8 canonical works of Brazilian literature; (3) a multi-dimensional 
assessment framework with difficulty levels, multi-turn evaluation, and 
optional coherence scoring; and (4) a comprehensive evaluation suite with 500 prompts: 200 single-turn prompts and 100 multi-turn dialogues with three turns each (300 turns total), baselines for 18 models, and open-source code.

\section{Related Work}
\label{sec:related}

\subsection{Instruction-Following Evaluation}
The systematic evaluation of instruction-following capabilities emerged as LLMs became increasingly deployed in applications requiring precise output control. \citet{zhou2023instruction} introduced IFEval, establishing a paradigm of verifiable instructions with constraints that can be automatically checked without subjective judgment. Their 25 instruction types cover word counts, formatting requirements, keyword inclusion, and structural constraints, evaluated on 541 English prompts.

Building on this foundation, \citet{jiang2024followbench} proposed FollowBench, which uses a multi-level mechanism to incrementally add constraints, creating five levels (1–5 constraints) across five constraint categories, evaluated in single-round settings over 820 curated instructions. \citet{qin2024infobench} contributed InFoBench with 500 prompts, focusing on decomposed instruction evaluation to assess granular capabilities and identify specific failure modes in instruction-following.

\citet{pyatkin2025ifbench} extended this paradigm with IFBench, introducing 58 new verifiable out-of-domain constraint templates paired with 300 prompts to test generalization to unseen constraint formulations. They show models often overfit prior constraint sets and struggle on novel constraints.

Multi-IF~\citep{he2024multiif} addressed the multilingual dimension by translating IFEval's 25 instruction types to seven languages (Chinese, French, Hindi, Italian, Portuguese, Russian, and Spanish) plus English, creating a dataset of 3,192 prompts. The benchmark extends evaluation to multi-turn conversations with three turns, testing whether models maintain instruction adherence across dialogue. While Multi-IF represents an important milestone in multilingual instruction-following evaluation, its translation-based approach has inherent limitations: the 25 instruction types are designed for English and may not capture language-specific phenomena, prompts retain English-centric contexts and cultural references, and the evaluation cannot assess linguistic features unique to target languages. For Portuguese specifically, Multi-IF does not evaluate constraints related to word termination patterns, Portuguese discourse markers, or culturally grounded content, gaps that CAPITU addresses through native Portuguese instruction design.

\subsection{Native and Culturally-Grounded Benchmarks}
The NLP community has begun to shift focus from broad multilingual understanding to the evaluation of native instruction-following capabilities, recognizing that translation-based constraints often fail to capture the complexity of specific languages. Recent benchmarks have emerged to address this by incorporating constraints deeply rooted in local linguistic and cultural norms. For instance, CIF-Bench~\citep{yang2024cif} evaluates how well LLMs generalize to Chinese through a large set of tasks developed with native-speaker input across diverse categories, including culturally grounded Chinese tasks and language-specific phenomena such as Chinese idioms, going beyond purely generic formatting constraints.

Similarly, OuLiBench~\citep{oulibench2025} introduces formal literary constraints inspired by the Oulipo movement (e.g., lipograms and tautograms) to assess linguistic competence in Italian, using rigorous, language-intrinsic restrictions as a lens for probing controllable generation. 

These initiatives highlight a critical evolution in evaluation methodologies: the move towards verifying whether models can satisfy structural and stylistic constraints that are native to their training data's culture. However, despite these advancements for Chinese and Italian, a comparable framework that intertwines the rich literary and normative context of Portuguese with verifiable structural constraints remains unexplored.

\subsection{Instruction-Tuning and Evaluation in Portuguese}
The landscape of Portuguese NLP has shifted rapidly from semantic similarity tasks to generative instruction-tuning. More recently, the focus has moved toward robust, native evaluation suites. PoETa v2 \citep{almeida2025poeta} introduces a 44-task benchmark for Portuguese, including 12 natively designed tasks grounded in regional knowledge (e.g., local exams, proverbs, and colloquial expressions) and 32 translated tasks adapted from established English benchmarks. Although PoETa v2 is not an instruction-following benchmark in the verifiable-constraint sense, it still provides an indirect signal of instruction adherence in Portuguese: many tasks require strict, automatically checkable response schemas (e.g., selecting a multiple-choice option, producing a binary answer, assigning a label, or extracting an answer span), and deviations from the expected format directly translate into errors. 

Similarly, the Sabiá-4 technical report \cite{laitz2026sabi} highlights the use of domain-specific instruction benchmarks, such as the Brazilian Bar Exam (OAB), where models must follow rigid structural guidelines to draft legal documents.

However, neither PoETa v2 nor OAB-based evaluations systematically compose and vary explicit constraints (e.g., length limits, keyword inclusion, formatting templates, or generalization to unseen constraint formulations), and therefore they cannot isolate instruction-following failure modes, motivating native Portuguese instruction-following benchmarks with explicitly verifiable constraints.

Moreover, another critical distinction remains: these benchmarks primarily assess knowledge retention and reasoning capabilities rather than the mechanical ability to strictly follow verifiable constraints. The current ecosystem relies on implicit instruction-following embedded in exams, leaving a gap for a dedicated benchmark like CAPITU that isolates and evaluates the steerability of models using native linguistic and literary constraints.

\subsection{Positioning of CAPITU}
CAPITU addresses three critical gaps simultaneously: (1) the absence of Portuguese from instruction-following benchmarks, despite its global linguistic significance; (2) the lack of culturally-grounded evaluation that tests models on content central to Brazilian cultural identity; and (3) the need for multi-turn instruction-following assessment in non-English languages. By combining verifiable constraints from the IFEval paradigm with literary texts from the Brazilian canon, CAPITU provides a rigorous benchmark that evaluates both technical instruction-following capabilities and cultural-linguistic competence. Table~\ref{tab:related_compact} summarizes how CAPITU relates to existing benchmarks across these dimensions.

\begin{table*}[t]
\centering
\caption{Instruction-following benchmarks comparison: scope and evaluation characteristics.}
\label{tab:related_compact}
\small
\begin{tabular}{llccccc}
\toprule
\textbf{Category} & \textbf{Benchmark} & \textbf{Year} & \textbf{PT-BR} & \textbf{Instruction Types} & \textbf{Multi-turn} & \textbf{Cultural} \\
\midrule
\multirow{4}{*}{English IF}
& IFEval & 2023 & \xmark & 25 & \xmark & \xmark \\
& FollowBench & 2024 & \xmark & 5 levels & \cmark & \xmark \\
& IFBench & 2024 & \xmark & 58 & \xmark & \xmark \\
& InFoBench & 2024 & \xmark & Decomposed & \xmark & \xmark \\
\midrule
Multilingual IF
& Multi-IF & 2024 & \cmark$^\dagger$ & 25 & \cmark & \xmark  \\
\midrule
\multirow{2}{*}{Native Cultural}
& CIF-Bench (ZH) & 2024 & \xmark & Cultural & \xmark & \cmark  \\
& OuLiBench (IT) & 2025 & \xmark & Literary & \xmark & \cmark  \\
\midrule
\multirow{1}{*}{Portuguese NLP}
& PoETa v2 & 2024 & \cmark & - & \xmark & \cmark \\
\midrule
& \textbf{CAPITU (Ours)} & \textbf{2026} & \textbf{\cmark} & \textbf{59} & \textbf{\cmark} & \textbf{\cmark}  \\
\bottomrule
\end{tabular}
\begin{threeparttable}
\begin{tablenotes}
\footnotesize
\item IF = Instruction-Following; PT-BR = Brazilian Portuguese
\item $^\dagger$ Multi-IF includes Portuguese via translation, not native design.
\end{tablenotes}
\end{threeparttable}
\end{table*}

\section{Methodology}
\label{sec:methodology}

\begin{figure}[ht]
\centering
\includegraphics[width=\textwidth]{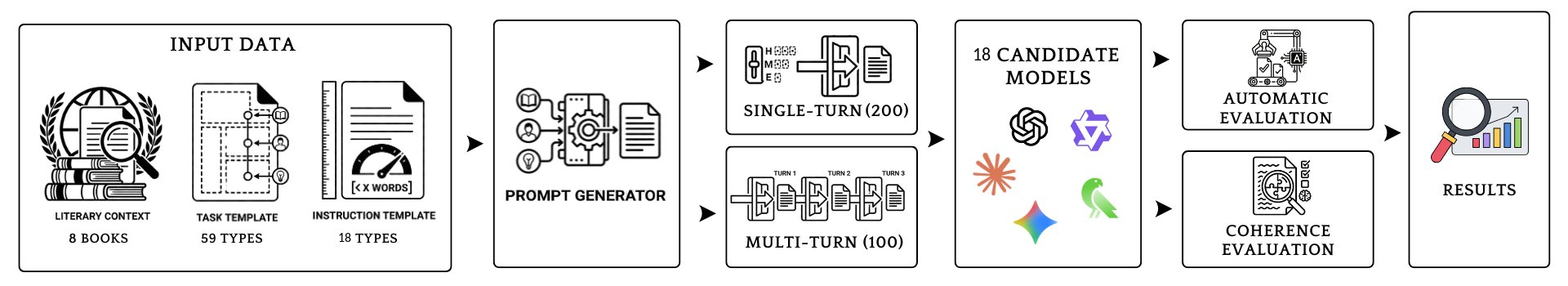}
\caption{Overview of the CAPITU methodology}
\label{fig:methodology}
\end{figure}

\subsection{Design Principles}

The design of CAPITU is guided by four foundational principles. The first and most critical principle is \textbf{automatic verification}: every instruction must be verifiable through programmatic methods such as string operations, regular expressions, or lexicon matching, without requiring human annotators or LLM judges. We deliberately exclude instructions that would require subjective judgment, such as ``use formal register'' or ``avoid clichés,'' because such constraints depend on open-ended lexicons that may not capture all relevant cases. This design choice enables fully automated evaluation at scale while avoiding false positives where a model might use an unlisted item and incorrectly pass verification.

The second principle, \textbf{linguistic relevance}, ensures that instructions reflect meaningful Portuguese linguistic phenomena rather than serving as mere translations of English constraints. Portuguese exhibits productive morphological patterns (like diminutives, gerunds and adverbial formations) that have no direct English equivalents and require native instruction design to evaluate properly. The third principle, \textbf{cultural grounding}, situates all prompts within Brazilian literary contexts, providing natural and coherent scenarios for text generation tasks. This grounding serves as context rather than evaluation target. Finally, \textbf{scalable difficulty} ensures that the benchmark supports evaluation across multiple complexity levels through two complementary mechanisms: (1) \textit{instruction composition}, which varies the number of simultaneous constraints from one (easy) to two (medium) to three or four (hard); and (2) \textit{multi-turn conversations}, where constraints accumulate across three turns, testing whether models maintain compliance as the conversation progresses.

\subsection{Instruction Taxonomy}

We define 59 instruction types organized into seven categories, each targeting different aspects of text generation control. Table~\ref{tab:instruction_categories} provides an overview of these categories with representative examples.

\begin{table}[ht]
\centering
\caption{Instruction categories in CAPITU with representative examples.}
\label{tab:instruction_categories}
\begin{tabular}{@{}llp{5.5cm}@{}}
\toprule
\textbf{Category} & \textbf{Count} & \textbf{Example Instructions} \\
\midrule
Count & 17 & Word count range, exact word count, sentence count, paragraph count, character count \\
Words & 11 & Include word, word frequency, conjunctions, connectives, temporal markers \\
Pattern & 8 & Words ending in -ando/-endo/-indo, -inho/-inha, -ão/-ões, -mente \\
Forbidden & 7 & Forbidden word, no numbers, no first person, no questions \\
Structure & 6 & Start with word, end with word, acrostic, line prefix \\
Punctuation & 5 & Only declarative, include question, use semicolon \\
Format & 5 & Bullet list, numbered list, all caps, all lowercase \\
\midrule
\textbf{Total} & \textbf{59} & \\
\bottomrule
\end{tabular}
\end{table}

\subsubsection{Portuguese-Specific Instructions}

Several instruction types are designed specifically to capture Portuguese linguistic phenomena that cannot be meaningfully assessed through translated English constraints. The most distinctive category involves word termination patterns, which exploit Portuguese's productive morphological system. We include constraints for words ending in \textit{-ando}, \textit{-endo}, and \textit{-indo} (common in gerund constructions), \textit{-inho} and \textit{-inha} along with their variants \textit{-zinho} and \textit{-zinha} (diminutive suffixes), \textit{-ão} and \textit{-ões} (augmentative and plural patterns), and \textit{-mente} (adverb formation). For each pattern, we define instructions requiring minimum usage, maximum limits, or complete prohibition, enabling fine-grained assessment of morphological control.

Importantly, these instructions test the model's ability to control surface-level morphological patterns, not the semantically correct use of the underlying grammatical category. For instance, the constraint ``include at least 3 words ending in -inho/-inha'' does not require that these words be true diminutives (any word matching the suffix pattern is accepted), consistent with our principle of automatic verification via deterministic string matching. The goal is to force models to handle productive morphological phenomena typical of Portuguese that English-centric benchmarks leave untested, such as diminutive and augmentative suffixes, gerund forms, and adverbial formations.

Beyond morphological patterns, we include constraints targeting discourse markers that are essential for textual cohesion in Portuguese. These comprise connectives (\textit{portanto}, \textit{contudo}, \textit{todavia}, \textit{além disso}), temporal markers (\textit{primeiro}, \textit{depois}, \textit{por fim}, \textit{finalmente}), and contrast markers (\textit{por outro lado}, \textit{em contrapartida}, \textit{no entanto}). We use closed lexicons containing only unambiguous markers to ensure deterministic verification. Similarly, constraints on grammatical person employ comprehensive Portuguese pronoun lists that include all variations (\textit{eu}, \textit{me}, \textit{mim}, \textit{comigo}, \textit{meu}, \textit{minha}, \textit{nós}, \textit{nosso}, and their equivalents for other persons), enabling precise control over narrative voice.

\subsubsection{Structural Instructions}

Complementing the linguistic constraints, we include structural instructions that test precise text formatting capabilities. The most challenging is the acrostic constraint, which requires the first letters of each sentence to form a specified word such as AMOR, VIDA, or ARTE. This instruction demands careful planning and simultaneous attention to content coherence. Other structural constraints include line prefix requirements (each line must begin with a specific character), prohibitions on repeating sentence-initial words in consecutive sentences, and requirements that responses begin or end with particular words.

\subsection{Literary Corpus}

All prompts in CAPITU are contextualized within eight canonical works of Brazilian literature, selected to represent major literary movements and provide diverse linguistic challenges. Table~\ref{tab:literary_works} summarizes the corpus.

\begin{table}[ht]
\centering
\caption{Literary works included in CAPITU.}
\label{tab:literary_works}
\begin{tabular}{@{}llll@{}}
\toprule
\textbf{Title} & \textbf{Author} & \textbf{Year} & \textbf{Movement} \\
\midrule
Dom Casmurro & Machado de Assis & 1899 & Realism \\
O Cortiço & Aluísio Azevedo & 1890 & Naturalism \\
Iracema & José de Alencar & 1865 & Romanticism \\
Grande Sertão: Veredas & Guimarães Rosa & 1956 & Modernism \\
Macunaíma & Mário de Andrade & 1928 & Modernism \\
Vidas Secas & Graciliano Ramos & 1938 & Regionalism \\
Capitães da Areia & Jorge Amado & 1937 & Modernism \\
A Hora da Estrela & Clarice Lispector & 1977 & Modernism \\
\bottomrule
\end{tabular}
\end{table}

The selection criteria prioritize cultural significance, diversity of narrative styles, and public domain availability. These works span over a century of Brazilian literary production and encompass markedly different linguistic registers, from Machado de Assis's ironic, urbane prose to Guimarães Rosa's experimental synthesis of regional speech and neologism. This diversity ensures that models must demonstrate robust Portuguese comprehension rather than relying on superficial pattern matching. We provide structured metadata for each work, including characters, themes, settings, narrator type, and historical context, in a companion file used for optional coherence evaluation.

\subsection{Prompt Generation Pipeline}

CAPITU employs a fully programmatic pipeline for prompt generation, ensuring reproducibility, balanced coverage, and systematic variation across literary works, difficulty levels, instruction categories, and task types. The pipeline operates without any LLM involvement in prompt creation, relying instead on template-based generation with parameterized randomization.

\subsubsection{Architecture Overview}

The generation system consists of three main components: (1) instruction templates that define constraint types with parameterized descriptions; (2) task templates that provide literary-contextualized prompts; and (3) a composition engine that combines these elements according to difficulty-specific rules. Each instruction template includes an identifier, a natural language description template with placeholders, a parameter generator function, and a category label. For example, the word count range instruction has the template ``a resposta deve conter entre \{min\_words\} e \{max\_words\} palavras'' with a generator that produces values like $\{$min\_words: 120, max\_words: 150$\}$.

\paragraph{Task Templates}

Task templates ground all prompts in Brazilian literary contexts. We define 13 distinct task formulations that span common literary analysis activities:

\begin{itemize}
    \item \textbf{Critical analysis}: ``Escreva uma análise crítica de `\{title\}' de \{author\}, explorando como a obra aborda \{theme\}.''
    \item \textbf{Reading club invitation}: ``Escreva uma carta para um clube de leitura convidando o grupo a debater `\{title\}', destacando \{theme\}.''
    \item \textbf{Book review}: ``Produza uma breve resenha crítica de `\{title\}', analisando o tratamento de \{theme\}.''
    \item \textbf{Comparative discussion}: ``Compare a abordagem de \{theme\} em `\{title\}' com outras obras de \{author\}.''
\end{itemize}

\paragraph{Instruction Parameter Generation}

Each instruction type includes a parameter generator that produces constraint values within appropriate ranges. For count-based instructions, generators sample from predefined value sets: word count ranges draw minimum values from $\{80, 100, 120, 150, 180\}$ and add random offsets of 20--50 words for maximums; sentence counts sample from $\{5, 6, 7, 8, 10\}$; paragraph counts from $\{2, 3, 4\}$. For word-based instructions, generators sample from curated lexicons: forbidden words draw from common filler words like ``muito'', ``sempre'', ``realmente'', ``basicamente''; required words draw from literary vocabulary like ``narrativa'', ``personagem'', ``identidade'', ``memória''. For structural instructions, acrostic words sample from $\{$AMOR, VIDA, ARTE, OBRA, LER, SOL, MAR, CÉU$\}$, and line prefixes from $\{$•, -, *, >, →$\}$.

\paragraph{Difficulty-Based Composition}

We define three difficulty levels based on the number of simultaneous constraints, as summarized in Table~\ref{tab:difficulty_levels}.

\begin{table}[ht]
\centering
\caption{Difficulty levels and constraint composition.}
\label{tab:difficulty_levels}
\small
\begin{tabular}{@{}lccp{5.5cm}@{}}
\toprule
\textbf{Level} & \textbf{Instructions} & \textbf{Pool Size} & \textbf{Example Combination} \\
\midrule
Easy & 1 & 14 & word\_count\_range \\
Medium & 2 & 25 pairs & word\_count\_range + no\_questions \\
Hard & 3--4 & 23 combos & exact\_word\_count + no\_questions + terminacao\_mente\_limit \\
\bottomrule
\end{tabular}
\end{table}

\textbf{Easy} questions use single constraints drawn from basic counts, simple prohibitions, and formatting rules. \textbf{Medium} questions pair constraints from different categories (e.g., a count constraint with a prohibition). \textbf{Hard} questions combine three or four constraints that create competing demands---for instance, maintaining an exact word count while incorporating required morphological patterns and avoiding certain punctuation.

All combinations are manually curated to ensure logical consistency; contradictory pairs (e.g., \texttt{include\_question} + \texttt{no\_questions}) are excluded. The complete list of combinations is provided in Appendix~\ref{app:combinations}.

\paragraph{Prompt Assembly}

The final prompt assembles task and instruction components using a standardized format. First, the task template is instantiated with a randomly selected book, author, and theme. Then, instruction descriptions are concatenated with the prefix ``Siga estritamente:'' (Follow strictly:) and joined with semicolons. Figure~\ref{fig:prompt_example} illustrates the structure.

\begin{figure}[ht]
\centering
\begin{tcolorbox}[
    colback=gray!5,
    colframe=gray!50,
    width=0.95\textwidth,
    arc=2mm,
    boxrule=0.5pt,
    title={\small\textbf{Example Prompt (Medium Difficulty)}},
    fonttitle=\sffamily
]
\small
\textit{Escreva uma análise crítica de} \textbf{`Dom Casmurro'} \textit{de} \textbf{Machado de Assis}\textit{, explorando como a obra aborda} \textbf{identidade e pertencimento}\textit{.}

\vspace{0.5em}
\textbf{Siga estritamente:} resposta deve conter entre \textbf{120} e \textbf{150} palavras; use no máximo \textbf{5} palavras terminadas em \textbf{-mente}
\end{tcolorbox}
\caption{Example of an assembled prompt showing task template (top) and instruction constraints (bottom).}
\label{fig:prompt_example}
\end{figure}

\subsection{Multi-Turn Evaluation}

Following the methodology established by Multi-IF~\citep{he2024multiif}, we support multi-turn conversations where instructions accumulate across turns. Each conversation comprises three turns with progressively accumulating constraints, testing models' ability to maintain constraint satisfaction across conversation history, a capability essential for real-world applications involving iterative refinement.
\subsubsection{Conversation Templates}

We define three conversation types to capture different analytical trajectories. Each type progresses through three turns with increasing analytical depth.

\begin{figure}[ht]
\centering
\begin{tcolorbox}[
    colback=blue!3,
    colframe=blue!40,
    width=0.95\textwidth,
    arc=2mm,
    boxrule=0.5pt,
    title={\small\textbf{Expansion} — Broadens analysis scope},
    fonttitle=\sffamily
]
\small
\textbf{Turn 1:} Apresente brevemente a obra `\{title\}' de \{author\}.\\
\textbf{Turn 2:} Agora aprofunde a análise, focando em \{theme\}.\\
\textbf{Turn 3:} Para concluir, compare esta obra com outra do mesmo período literário.
\end{tcolorbox}

\vspace{0.3em}

\begin{tcolorbox}[
    colback=green!3,
    colframe=green!40,
    width=0.95\textwidth,
    arc=2mm,
    boxrule=0.5pt,
    title={\small\textbf{Deepening} — Focuses on specific aspects},
    fonttitle=\sffamily
]
\small
\textbf{Turn 1:} Quem é o protagonista de `\{title\}'? Descreva suas características.\\
\textbf{Turn 2:} Como esse personagem se relaciona com o tema de \{theme\}?\\
\textbf{Turn 3:} Analise a evolução desse personagem ao longo da narrativa.
\end{tcolorbox}

\vspace{0.3em}

\begin{tcolorbox}[
    colback=orange!3,
    colframe=orange!40,
    width=0.95\textwidth,
    arc=2mm,
    boxrule=0.5pt,
    title={\small\textbf{Critical} — Develops argumentative analysis},
    fonttitle=\sffamily
]
\small
\textbf{Turn 1:} Identifique o principal recurso literário usado em `\{title\}'.\\
\textbf{Turn 2:} Explique como esse recurso contribui para a representação de \{theme\}.\\
\textbf{Turn 3:} Apresente uma análise crítica sobre a eficácia desse recurso.
\end{tcolorbox}
\end{figure}

\subsubsection{Progressive Instruction Accumulation}

Instructions are added progressively across turns, with each turn introducing new constraints while maintaining all previous ones.
At each turn, the prompt includes all accumulated instruction descriptions. By Turn 3, responses must satisfy three simultaneous constraints while maintaining coherence with the evolving analytical focus.

\subsection{Evaluation Framework}

\subsubsection{Instruction Verification}

Each of the 59 instruction types has a corresponding verification function implemented using deterministic methods. We organize verification into four categories based on implementation approach: (1) count-based verification using string operations for word, sentence, paragraph, and character counting; (2) pattern-based verification using regular expressions for Portuguese morphological constraints; (3) lexicon-based verification matching against closed lists of pronouns, connectives, and discourse markers; and (4) structural verification for text organization, delimiters, and positional constraints.

The verification functions are strictly objective and rule-based. By relying on text-matching and logical constraints rather than external APIs or model-based judgment, the framework eliminates randomness and ensures identical results across evaluation runs. Implementation details, including tokenization rules, regular expression patterns, and complete lexicons, are provided in Appendix~\ref{app:verification_impl}.

\subsubsection{Metrics}

We report multiple accuracy metrics to capture different aspects of instruction-following performance. Strict accuracy measures the percentage of prompts where all instructions are satisfied on the original response exactly as generated. Loose accuracy applies the same criterion but tests each instruction against multiple response variants: the original response, versions with the first or last line removed, and versions with markdown formatting stripped. An instruction passes under loose evaluation if any variant satisfies the constraint. This accommodates models that prepend markdown headers or append formatting artifacts that cause strict failures on otherwise compliant responses. Instruction-level accuracy measures the percentage of individual instructions satisfied across all prompts, providing insight into which constraint types prove most challenging.

For multi-turn evaluation, we additionally report turn-wise accuracy, which tracks performance at each conversation turn, and conversation accuracy, which measures the percentage of conversations where all turns satisfy all applicable constraints. These metrics reveal how performance degrades as constraints accumulate.

\subsubsection{Coherence Evaluation}

A fundamental limitation of purely constraint-based evaluation is that it cannot distinguish between substantive responses and degenerate solutions that technically satisfy instructions without producing meaningful content. Consider a prompt requesting ``an 80-word summary of Dom Casmurro'': a model could repeat the word ``resumo'' (summary) 80 times and achieve perfect instruction compliance, receiving the same score as a model that produces a genuine, coherent summary of the novel. Similarly, a model could satisfy a ``use at least 3 words ending in -mente'' constraint by producing ``realmente realmente realmente'' without any contextual relevance.

This vulnerability is not hypothetical, instruction-following benchmarks that rely solely on mechanical verification can incentivize models to optimize for constraint satisfaction at the expense of response quality. To address this limitation, CAPITU incorporates optional coherence evaluation that penalizes degenerate responses and rewards substantive content that genuinely engages with the literary prompts.

\paragraph{Coherence Assessment}

To assess overall coherence and clarity, we utilize an LLM-based judge. The model is prompted to evaluate the responses specifically in the context of Brazilian Portuguese, relying strictly on the provided literary metadata to avoid the influence of external knowledge. The evaluation results in a coherence score ranging from 0 to 1, accompanied by a brief qualitative justification.

The final combined score integrates instruction compliance ($I$) and coherence ($C$) with a 2:1 weighting:

\begin{equation}
\text{Final Score} = \frac{2 \cdot I + C}{3}
\label{eq:final_score}
\end{equation}

The 2:1 weighting reflects our design priority: CAPITU primarily evaluates instruction-following, with coherence serving as a safeguard against degenerate solutions. Alternative weightings are supported in our evaluation code. This weighting prioritizes constraint satisfaction (the primary evaluation target) while ensuring that degenerate solutions receive substantially lower scores than coherent responses. A model that achieves 100\% instruction compliance but produces incoherent text would score 0.67, whereas a model with both perfect compliance and coherence scores 1.0. This design ensures that gaming the constraints through repetition or irrelevant content is penalized in the final evaluation.

\section{Benchmark Statistics}
\label{sec:statistics}

Table~\ref{tab:benchmark_stats} summarizes CAPITU's composition. The benchmark totals 500 evaluation instances: 200 single-turn questions across three difficulty levels and 100 multi-turn conversations with three turns each (resulting in 300 evaluation instances).

\begin{table}[ht]
\centering
\caption{CAPITU benchmark statistics.}
\label{tab:benchmark_stats}
\small
\renewcommand{\arraystretch}{1.1}
\begin{tabular}{@{}lr@{\hskip 3em}lr@{\hskip 3em}lr@{}}
\toprule
\multicolumn{2}{c}{\textbf{Benchmark Design}} & \multicolumn{2}{c}{\textbf{Single-turn}} & \multicolumn{2}{c}{\textbf{Multi-turn}} \\
\midrule
Instruction types & 59 & Easy & 70 & Conversations & 100 \\
Instruction categories & 7 & Medium & 70 & Turns each & 3 \\
Literary works & 8 & Hard & 60 & \textbf{Total turns} & \textbf{300} \\
Literary movements & 5 & \textbf{Total} & \textbf{200} & & \\
Difficulty levels & 3 & & & & \\
\bottomrule
\end{tabular}
\end{table}

\section{Experiments}
\label{sec:experiments}

\subsection{Experimental Setup}

We evaluate 18 state-of-the-art LLMs on CAPITU to establish baseline performance and analyze instruction-following capabilities in Portuguese. The evaluated models are listed below. Models marked with $^\dagger$ feature native reasoning capabilities.
\begin{itemize}
    \item \textbf{OpenAI GPT-5 family}~\citep{singh2025openai}: GPT-5-mini$^\dagger$, GPT-5$^\dagger$, GPT-5.2\footnote{Default configuration, no reasoning tokens used.}, GPT-5.2 (reasoning-effort high)$^\dagger$\footnote{Configured with \texttt{reasoning\_effort=high}, enabling extended reasoning tokens.}
    \item \textbf{Google Gemini 3}~\citep{pichai2025new}: Gemini-3-Pro-Preview$^\dagger$, Gemini-3-Flash-Preview
    \item \textbf{Anthropic Claude 4.5 \citealp{anthropic2025claude45}}: Claude-Sonnet-4.5, Claude-Haiku-4.5
    \item \textbf{Alibaba Qwen 3}~\citep{yang2025qwen3}: Qwen3-8B$^\dagger$, Qwen3-14B$^\dagger$, Qwen3-32B$^\dagger$, Qwen3-235b-a22b, Qwen3-235b-a22b-thinking$^\dagger$
    \item \textbf{Maritaca AI Sabiá}~\citep{abonizio2024sabi} (Portuguese-specialized): Sabiá-3, Sabiazinho-3, Sabiá-3.1, Sabiá-4, Sabiazinho-4
\end{itemize}

\subsubsection{Evaluation Protocol}

For each model, we generate responses using temperature 0.0 for reproducibility, evaluate on 200 single-turn prompts and 100 multi-turn conversations (3 turns each), and apply both strict verification (exact response match) and loose verification (testing response variants with first/last line removal and markdown stripping to accommodate formatting artifacts). We also compute coherence scores using GPT-4o-mini as judge with access to literary context, record API costs for cost-performance analysis, and report metrics across all breakdown dimensions (difficulty, category, literary work, turn).

\section{Results}
\label{sec:results}

We evaluate 18 models across three categories: proprietary models (OpenAI GPT-5 family, Google Gemini 3, Anthropic Claude 4.5), open-weight models (Alibaba Qwen 3 series), and Portuguese-specialized proprietary models (Maritaca AI Sabiá family).

\subsection{Overall Performance}

Table~\ref{tab:main_results} presents single-turn evaluation results across all models, sorted by final score.

\small
\setlength{\tabcolsep}{12pt}
\renewcommand{\arraystretch}{1.2}

\begin{longtable}{@{}c S S S S S@{}}
\caption{Single-turn results on CAPITU. \textbf{Strict}/\textbf{Loose} = prompt-level accuracy (all instructions correct; Loose tests response variants with first/last line removal and markdown stripping); \textbf{Instr.} = instruction-level accuracy; \textbf{Coh.} = coherence score (GPT-4o-mini judge); \textbf{Final} = average of per-prompt scores computed via Equation~\eqref{eq:final_score}. Cell backgrounds indicate performance.} 
\label{tab:main_results} \\

\toprule
\textbf{Model} & {\textbf{Strict}} & {\textbf{Loose}} & {\textbf{Inst.}} & {\textbf{Coh.}} & {\textbf{Final}} \\
\midrule
\endfirsthead

\toprule
\textbf{Model} & {\textbf{Strict}} & {\textbf{Loose}} & {\textbf{Inst.}} & {\textbf{Coh.}} & {\textbf{Final}} \\
\midrule
\endhead

\bottomrule
\endfoot

\multicolumn{6}{@{}l}{\textit{Proprietary Models}} \\[2pt]

GPT-5$^\dagger$
  & \cellcolor{perf6}98.0
  & \cellcolor{perfTop}\textbf{99.0}
  & \cellcolor{perfTop}99.0
  & \cellcolor{coh4}\textbf{99.0}
  & \cellcolor{perfTop}\textbf{99.1} \\

GPT-5-mini$^\dagger$
  & \cellcolor{perf6}98.0
  & \cellcolor{perf6}98.0
  & \cellcolor{perfTop}99.0
  & \cellcolor{coh3}98.1
  & \cellcolor{perf6}98.9 \\

GPT-5.2$^\dagger$
  & \cellcolor{perf6}\textbf{98.5}
  & \cellcolor{perf6}98.5
  & \cellcolor{perfTop}\textbf{99.3}
  & \cellcolor{coh3}97.6
  & \cellcolor{perf6}98.8 \\

GPT-5.2
  & \cellcolor{perf5}91.5
  & \cellcolor{perf5}91.5
  & \cellcolor{perf6}95.1
  & \cellcolor{coh3}98.4
  & \cellcolor{perf6}97.1 \\

Gemini-3-Pro-Preview$^\dagger$
  & \cellcolor{perf5}92.5
  & \cellcolor{perf6}95.5
  & \cellcolor{perf6}95.3
  & \cellcolor{coh3}98.7
  & \cellcolor{perf6}97.1 \\

Gemini-3-Flash-Preview
  & \cellcolor{perf4}89.0
  & \cellcolor{perf5}93.0
  & \cellcolor{perf5}94.3
  & \cellcolor{coh3}97.7
  & \cellcolor{perf6}96.7 \\

Claude-Sonnet-4.5
  & \cellcolor{perf4}86.0
  & \cellcolor{perf5}93.0
  & \cellcolor{perf5}90.6
  & \cellcolor{coh4}98.8
  & \cellcolor{perf5}94.3 \\

Claude-Haiku-4.5
  & \cellcolor{perf1}73.5
  & \cellcolor{perf3}84.0
  & \cellcolor{perf3}82.5
  & \cellcolor{coh3}98.3
  & \cellcolor{perf4}89.2 \\

\midrule

\multicolumn{6}{@{}l}{\textit{Open-Weight Models}} \\[2pt]

Qwen3-8B$^\dagger$
  & \cellcolor{perf3}81.5
  & \cellcolor{perf4}85.0
  & \cellcolor{perf5}90.4
  & \cellcolor{coh1}92.2
  & \cellcolor{perf5}92.7 \\

Qwen3-32B$^\dagger$
  & \cellcolor{perf3}82.5
  & \cellcolor{perf3}83.5
  & \cellcolor{perf5}90.1
  & \cellcolor{coh2}96.2
  & \cellcolor{perf5}92.6 \\

Qwen3-14B$^\dagger$
  & \cellcolor{perf3}81.0
  & \cellcolor{perf3}83.0
  & \cellcolor{perf4}89.4
  & \cellcolor{coh1}92.7
  & \cellcolor{perf5}90.8 \\

Qwen3-235b-a22b
  & \cellcolor{perf2}79.5
  & \cellcolor{perf3}84.5
  & \cellcolor{perf4}89.4
  & \cellcolor{coh2}96.8
  & \cellcolor{perf5}91.7 \\

Qwen3-235b-a22b$^\dagger$
  & \cellcolor{perf3}84.0
  & \cellcolor{perf3}84.0
  & \cellcolor{perf4}87.7
  & \cellcolor{coh2}96.2
  & \cellcolor{perf5}92.6 \\

\midrule

\multicolumn{6}{@{}l}{\textit{Portuguese-Specialized Proprietary Models}} \\[2pt]

Sabiazinho-4
  & \cellcolor{perf4}87.0
  & \cellcolor{perf4}87.0
  & \cellcolor{perf5}93.1
  & \cellcolor{coh2}94.1
  & \cellcolor{perf5}94.6 \\

Sabiá-4
  & \cellcolor{perf3}84.5
  & \cellcolor{perf3}84.5
  & \cellcolor{perf5}91.6
  & \cellcolor{coh2}96.0
  & \cellcolor{perf5}94.4 \\

Sabiá-3
  & \cellcolor{perf2}77.0
  & \cellcolor{perf3}81.0
  & \cellcolor{perf4}86.4
  & \cellcolor{coh3}97.0
  & \cellcolor{perf5}90.7 \\

Sabiazinho-3
  & \cellcolor{perf3}80.0
  & \cellcolor{perf3}80.0
  & \cellcolor{perf4}87.4
  & \cellcolor{coh2}91.7
  & \cellcolor{perf5}90.2 \\
  
Sabiá-3.1
  & \cellcolor{perf1}74.0
  & \cellcolor{perf2}79.5
  & \cellcolor{perf4}85.4
  & \cellcolor{coh2}93.7
  & \cellcolor{perf4}89.2 \\

\\[-6pt]
\multicolumn{6}{@{}l}{\footnotesize $^\dagger$ Reasoning/thinking enabled; GPT-5.2 uses \texttt{reasoning\_effort=high}.} \\

\end{longtable}
\noindent{\footnotesize Color scale: \colorbox{perf1}{\strut\small\,low\,}\;\colorbox{perf2}{\strut}\;\colorbox{perf3}{\strut}\;\colorbox{perf4}{\strut}\;\colorbox{perf5}{\strut}\;\colorbox{perf6}{\strut}\;\colorbox{perfTop}{\strut\small\,high\,}}\par\medskip

The OpenAI GPT-5 family dominates the benchmark: GPT-5.2 with reasoning-effort high achieves 98.5\% strict accuracy, the highest observed, followed closely by GPT-5 and GPT-5-mini, both at 98.0\%, substantially outperforming all other model families. Portuguese specialization also proves beneficial, as Sabiazinho-4 (87.0\%) surpasses Claude-Sonnet-4.5 (86.0\%) and Sabiá-4 (84.5\%) surpasses Qwen3-235b-a22b (79.5\%), the largest open-weight model evaluated in this study. Regarding model scale, smaller Qwen models with reasoning (8B at 81.5\%, 32B at 82.5\%) slightly outperform the larger 235B variant (79.5\%), suggesting that explicit reasoning capability matters more than raw parameter count for instruction-following. Finally, coherence remains high across the board: all models achieve scores above 0.91, indicating that instruction-following constraints do not significantly impair response quality. A detailed cost-performance analysis is provided in Appendix~\ref{app:cost}.

\subsection{Performance by Difficulty}

Table~\ref{tab:difficulty_results} breaks down strict accuracy by difficulty level (1, 2, or 3--4 simultaneous constraints).

\small
\setlength{\tabcolsep}{6pt}
\renewcommand{\arraystretch}{1.25}
\begin{longtable}{@{}cSSS|S@{}}
\caption{Strict accuracy (\%) by difficulty level. Easy = 1 instruction; Medium = 2 instructions; Hard = 3--4 instructions. Cell backgrounds indicate performance (see color scale below table).}
\label{tab:difficulty_results} \\
\toprule
\textbf{Model} & \textbf{Easy} & \textbf{Medium} & \textbf{Hard} & \textbf{$\Delta$ (E$\rightarrow$H)} \\
\midrule
\multicolumn{5}{@{}l}{\textit{Proprietary Models}} \\[2pt]

GPT-5$^\dagger$
  & \cellcolor{perfTop}\textbf{100.0}
  & \cellcolor{perf6}97.1
  & \cellcolor{perf6}\textbf{96.7}
  & \textbf{--3.3} \\

GPT-5-mini$^\dagger$
  & \cellcolor{perfTop}\textbf{100.0}
  & \cellcolor{perf6}\textbf{98.6}
  & \cellcolor{perf6}95.0
  & {--5.0} \\

GPT-5.2$^\dagger$		
	  & \cellcolor{perfTop}\textbf{100.0}	
	  & \cellcolor{perf6}\textbf{98.6}		
	  & \cellcolor{perf6}\textbf{96.7}		
	  & \textbf{--3.3} \\

GPT-5.2
  & \cellcolor{perfTop}\textbf{100.0}
  & \cellcolor{perf5}91.4
  & \cellcolor{perf3}81.7
  & {--18.3} \\

Gemini-3-Pro-Preview$^\dagger$
  & \cellcolor{perf6}95.7
  & \cellcolor{perf5}94.3
  & \cellcolor{perf4}86.7
  & {--9.0} \\

Gemini-3-Flash-Preview
  & \cellcolor{perfTop}\textbf{100.0}
  & \cellcolor{perf6}95.7
  & \cellcolor{perf1}68.3
  & {--31.7} \\

Claude-Sonnet-4.5
  & \cellcolor{perf5}92.9
  & \cellcolor{perf5}92.9
  & \cellcolor{perf1}70.0
  & {--22.9} \\

Claude-Haiku-4.5
  & \cellcolor{perf4}85.7
  & \cellcolor{perf3}84.3
  & \cellcolor{perf1}46.7
  & {--39.0} \\

\midrule

\multicolumn{5}{@{}l}{\textit{Open-Weight Models}} \\[2pt]

Qwen3-8B$^\dagger$
  & \cellcolor{perf6}97.1
  & \cellcolor{perf4}87.1
  & \cellcolor{perf1}56.7
  & {--40.4} \\

Qwen3-32B$^\dagger$
  & \cellcolor{perf5}91.4
  & \cellcolor{perf4}85.7
  & \cellcolor{perf1}68.3
  & {--23.1} \\

Qwen3-14B$^\dagger$
  & \cellcolor{perf5}91.4
  & \cellcolor{perf3}80.0
  & \cellcolor{perf1}70.0
  & {--21.4} \\

Qwen3-235b-a22b
  & \cellcolor{perf4}87.1
  & \cellcolor{perf3}84.3
  & \cellcolor{perf1}65.0
  & {--22.1} \\

Qwen3-235b-a22b$^\dagger$
  & \cellcolor{perf6}98.6
  & \cellcolor{perf4}85.7
  & \cellcolor{perf1}65.0
  & {--33.6} \\

\midrule

\multicolumn{5}{@{}l}{\textit{Portuguese-Specialized Proprietary Models}} \\[2pt]

Sabiazinho-4
  & \cellcolor{perfTop}\textbf{100.0}
  & \cellcolor{perf4}85.7
  & \cellcolor{perf1}73.3
  & {--26.7} \\

Sabiá-4
  & \cellcolor{perf6}98.6
  & \cellcolor{perf4}85.7
  & \cellcolor{perf1}66.7
  & {--31.9} \\

Sabiazinho-3
  & \cellcolor{perf5}94.3
  & \cellcolor{perf3}82.9
  & \cellcolor{perf1}60.0
  & {--34.3} \\

Sabiá-3
  & \cellcolor{perf4}88.6
  & \cellcolor{perf3}81.4
  & \cellcolor{perf1}58.3
  & {--30.3} \\

Sabiá-3.1
  & \cellcolor{perf5}90.0
  & \cellcolor{perf2}75.7
  & \cellcolor{perf1}53.3
  & {--36.7} \\

\bottomrule
\\[-6pt]
\multicolumn{5}{@{}l}{\footnotesize $^\dagger$ Reasoning/thinking enabled; GPT-5.2 uses \texttt{reasoning\_effort=high}.} \\
\end{longtable}
\noindent{\footnotesize Color scale: \colorbox{perf1}{\strut\small\,low\,}\;\colorbox{perf2}{\strut}\;\colorbox{perf3}{\strut}\;\colorbox{perf4}{\strut}\;\colorbox{perf5}{\strut}\;\colorbox{perf6}{\strut}\;\colorbox{perfTop}{\strut\small\,high\,}}\par\medskip

GPT-5 and GPT-5.2 (reasoning-effort high) demonstrate remarkable robustness, both with only a 3.3 percentage point drop between easy and hard questions---the smallest degradation observed. Most models show 20--40\% degradation on hard questions, highlighting the challenge of maintaining multiple simultaneous constraints. Qwen3-8B shows the largest drop (40.4 pp), despite achieving 97.1\% on easy questions, followed by Claude-Haiku-4.5 (39.0 pp) and Sabiá-3.1 (36.7 pp). Among open-weight models, Qwen3-235b-a22b with thinking enabled achieves 98.6\% on easy questions but drops to 65.0\% on hard, showing a 33.6 pp degradation. Notably, Gemini-3-Pro-Preview maintains the most balanced profile among non-GPT-5 models, with only a 9.0 pp drop.

\subsection{Multi-Turn Results}

Table~\ref{tab:multiturn_results} presents conversation-level and per-turn accuracy, testing constraint persistence across three turns with accumulating instructions.

\small
\setlength{\tabcolsep}{10pt}
\renewcommand{\arraystretch}{1.5}
\begin{longtable}{@{}c S S S S@{}}

\caption{Multi-turn evaluation. \textbf{Conver.} = conversation accuracy (all turns correct); \textbf{Turn} columns show per-turn strict accuracy with accumulating constraints. Cell backgrounds indicate performance (see color scale below table).}
\label{tab:multiturn_results} \\
\toprule
\textbf{Model} & \textbf{Conver.} & \textbf{Turn 1} & \textbf{Turn 2} & \textbf{Turn 3} \\
\midrule
\multicolumn{5}{@{}l}{\textit{Proprietary Models}} \\[2pt]

GPT-5-mini$^\dagger$
  & \cellcolor{perf6}\textbf{96.0}
  & \cellcolor{perfTop}\textbf{100.0}
  & \cellcolor{perf6}98.0
  & \cellcolor{perf6}\textbf{98.0} \\

Gemini-3-Pro-Preview$^\dagger$
  & \cellcolor{perf6}95.0
  & \cellcolor{perfTop}\textbf{100.0}
  & \cellcolor{perfTop}\textbf{99.0}
  & \cellcolor{perf6}96.0 \\

Gemini-3-Flash-Preview
  & \cellcolor{perf6}95.0
  & \cellcolor{perfTop}\textbf{100.0}
  & \cellcolor{perfTop}\textbf{99.0}
  & \cellcolor{perf6}95.0 \\

GPT-5$^\dagger$
  & \cellcolor{perf5}90.0
  & \cellcolor{perfTop}\textbf{100.0}
  & \cellcolor{perf5}91.0
  & \cellcolor{perf6}97.0 \\

Claude-Sonnet-4.5
  & \cellcolor{perf4}87.0
  & \cellcolor{perf6}96.0
  & \cellcolor{perf5}92.0
  & \cellcolor{perf4}88.0 \\

GPT-5.2
  & \cellcolor{perf1}73.0
  & \cellcolor{perf4}87.0
  & \cellcolor{perf3}83.0
  & \cellcolor{perf3}84.0 \\

GPT-5.2$^\dagger$		
  & \cellcolor{perf3}83.0		
  & \cellcolor{perf6}98.0		
  & \cellcolor{perf3}84.0		
  & \cellcolor{perf4}85.0 \\
  
Claude-Haiku-4.5
  & \cellcolor{perf1}72.0
  & \cellcolor{perf3}83.0
  & \cellcolor{perf3}84.0
  & \cellcolor{perf2}79.0 \\

\midrule

\multicolumn{5}{@{}l}{\textit{Open-Weight Models}} \\[2pt]

Qwen3-14B$^\dagger$
  & \cellcolor{perf1}74.0
  & \cellcolor{perf6}95.0
  & \cellcolor{perf5}90.0
  & \cellcolor{perf3}81.0 \\

Qwen3-8B$^\dagger$
  & \cellcolor{perf1}72.0
  & \cellcolor{perf6}95.0
  & \cellcolor{perf6}96.0
  & \cellcolor{perf1}74.0 \\

Qwen3-32B$^\dagger$
  & \cellcolor{perf1}60.0
  & \cellcolor{perf3}84.0
  & \cellcolor{perf3}84.0
  & \cellcolor{perf3}81.0 \\

Qwen3-235b-a22b
  & \cellcolor{perf1}60.0
  & \cellcolor{perf5}94.0
  & \cellcolor{perf3}83.0
  & \cellcolor{perf1}68.0 \\

Qwen3-235b-a22b$^\dagger$
  & \cellcolor{perf1}64.0
  & \cellcolor{perf5}92.0
  & \cellcolor{perf3}84.0
  & \cellcolor{perf1}68.0 \\

\midrule

\multicolumn{5}{@{}l}{\textit{Portuguese-Specialized Proprietary Models}} \\[2pt]

Sabiazinho-4
  & \cellcolor{perf4}87.0
  & \cellcolor{perf5}94.0
  & \cellcolor{perf6}98.0
  & \cellcolor{perf4}89.0 \\

Sabiá-4
  & \cellcolor{perf2}78.0
  & \cellcolor{perf4}88.0
  & \cellcolor{perf5}93.0
  & \cellcolor{perf4}85.0 \\

Sabiazinho-3
  & \cellcolor{perf1}68.0
  & \cellcolor{perf5}93.0
  & \cellcolor{perf5}90.0
  & \cellcolor{perf2}75.0 \\

Sabiá-3.1
  & \cellcolor{perf1}64.0
  & \cellcolor{perf6}97.0
  & \cellcolor{perf4}85.0
  & \cellcolor{perf1}70.0 \\

Sabiá-3
  & \cellcolor{perf1}64.0
  & \cellcolor{perf6}95.0
  & \cellcolor{perf4}86.0
  & \cellcolor{perf1}71.0 \\

\bottomrule
\multicolumn{5}{@{}l}{\footnotesize $^\dagger$ Model with reasoning/thinking enabled; GPT-5.2 uses \texttt{reasoning\_effort=high}.} \\
\end{longtable}
\noindent{\footnotesize Color scale: \colorbox{perf1}{\strut\small\,low\,}\;\colorbox{perf2}{\strut}\;\colorbox{perf3}{\strut}\;\colorbox{perf4}{\strut}\;\colorbox{perf5}{\strut}\;\colorbox{perf6}{\strut}\;\colorbox{perfTop}{\strut\small\,high\,}}\par\medskip
GPT-5-mini excels in multi-turn interactions (96.0\% conversation accuracy), outperforming GPT-5 (90.0\%), suggesting that reasoning overhead may occasionally hinder constraint tracking. Gemini models show strong turn consistency, maintaining 95--96\% accuracy even at Turn 3. Portuguese-specialized models show competitive multi-turn performance: Sabiazinho-4 achieves 87.0\% conversation accuracy, matching Claude-Sonnet-4.5, with strong constraint persistence. Among open-weight models, Qwen3-14B (74.0\%) and Qwen3-8B (72.0\%) show a sharp degradation pattern, with high Turn 1 accuracy (95\%) but significant drops at Turn 3. Qwen3-235b-a22b with thinking mode achieves 64.0\% conversation accuracy, with Turn 1 at 92.0\% dropping to 68.0\% at Turn 3, showing moderate constraint persistence degradation. Portuguese models degrade moderately: Sabiá-3.1 drops from 97\% (Turn 1) to 70\% (Turn 3), indicating difficulty maintaining the accumulated constraints.

\subsection{Performance by Instruction Category}

Table~\ref{tab:category_results} shows instruction-level accuracy by category for representative models across tiers.

\small
\setlength{\tabcolsep}{4pt}
\renewcommand{\arraystretch}{1.25}
\begin{xltabular}{\textwidth}{@{}Xccccccc@{}}
\caption{Instruction-level accuracy (\%) by category for representative models. Cell backgrounds indicate performance (see color scale below table).}
\label{tab:category_results} \\
\toprule
\textbf{Model} & \textbf{Count} & \textbf{Words} & \textbf{Pattern} & \textbf{Forbid.} & \textbf{Struct.} & \textbf{Punct.} & \textbf{Format} \\
\midrule
GPT-5.2$^\dagger$ & \cellcolor{perfTop}100.0 & \cellcolor{perfTop}100.0 & \cellcolor{perfTop}100.0 & \cellcolor{perfTop}100.0 & \cellcolor{perfTop}100.0 & \cellcolor{perf3}80.0 & \cellcolor{perf6}96.8 \\
GPT-5$^\dagger$ & \cellcolor{perfTop}99.2 & \cellcolor{perf6}98.2 & \cellcolor{perfTop}100.0 & \cellcolor{perfTop}100.0 & \cellcolor{perfTop}100.0 & \cellcolor{perf3}80.0 & \cellcolor{perfTop}100.0 \\
GPT-5-mini$^\dagger$ & \cellcolor{perfTop}99.2 & \cellcolor{perf6}98.2 & \cellcolor{perf6}98.6 & \cellcolor{perfTop}100.0 & \cellcolor{perfTop}100.0 & \cellcolor{perf5}90.0 & \cellcolor{perfTop}100.0 \\
Gemini-3-Pro-Preview$^\dagger$ & \cellcolor{perf6}94.9 & \cellcolor{perf6}98.2 & \cellcolor{perf5}94.3 & \cellcolor{perf6}97.7 & \cellcolor{perf4}87.5 & \cellcolor{perfTop}100.0 & \cellcolor{perf5}93.5 \\
Claude-Sonnet-4.5 & \cellcolor{perf4}87.3 & \cellcolor{perf6}98.2 & \cellcolor{perf4}87.1 & \cellcolor{perfTop}100.0 & \cellcolor{perf1}59.4 & \cellcolor{perfTop}100.0 & \cellcolor{perfTop}100.0 \\
Claude-Haiku-4.5 & \cellcolor{perf2}79.7 & \cellcolor{perf5}94.6 & \cellcolor{perf1}65.7 & \cellcolor{perfTop}100.0 & \cellcolor{perf1}37.5 & \cellcolor{perfTop}100.0 & \cellcolor{perfTop}100.0 \\
\midrule
Qwen3-235b-a22b$^\dagger$ & \cellcolor{perf3}82.2 & \cellcolor{perf5}91.1 & \cellcolor{perf3}84.3 & \cellcolor{perf6}95.5 & \cellcolor{perf6}96.9 & \cellcolor{perf5}90.0 & \cellcolor{perf2}77.4 \\
Qwen3-235b-a22b & \cellcolor{perf5}90.7 & \cellcolor{perf5}91.1 & \cellcolor{perf1}67.1 & \cellcolor{perfTop}98.9 & \cellcolor{perf6}96.9 & \cellcolor{perfTop}100.0 & \cellcolor{perf5}93.5 \\
\midrule
Sabiazinho-4 & \cellcolor{perf5}90.7 & \cellcolor{perf4}89.3 & \cellcolor{perf4}88.6 & \cellcolor{perfTop}98.9 & \cellcolor{perf6}96.9 & \cellcolor{perfTop}100.0 & \cellcolor{perf6}96.8 \\
Sabiazinho-3 & \cellcolor{perf3}83.1 & \cellcolor{perf3}83.9 & \cellcolor{perf2}78.6 & \cellcolor{perf6}97.7 & \cellcolor{perf4}87.5 & \cellcolor{perfTop}100.0 & \cellcolor{perf6}96.8 \\
Sabiá-3.1 & \cellcolor{perf3}82.2 & \cellcolor{perf4}89.3 & \cellcolor{perf1}72.9 & \cellcolor{perf6}97.7 & \cellcolor{perf2}75.0 & \cellcolor{perfTop}100.0 & \cellcolor{perf5}90.3 \\
\midrule
\textit{Avg. (18 models)} & \textit{89.4} & \textit{91.6} & \textit{84.8} & \textit{99.1} & \textit{87.9} & \textit{94.4} & \textit{93.4} \\
\bottomrule
\multicolumn{8}{@{}l}{\footnotesize $^\dagger$ Model with reasoning/thinking enabled; GPT-5.2 uses \texttt{reasoning\_effort=high}.} \\
\end{xltabular}
\noindent{\footnotesize Color scale: \colorbox{perf1}{\strut\small\,low\,}\;\colorbox{perf2}{\strut}\;\colorbox{perf3}{\strut}\;\colorbox{perf4}{\strut}\;\colorbox{perf5}{\strut}\;\colorbox{perf6}{\strut}\;\colorbox{perfTop}{\strut\small\,high\,}}\par\medskip

Several patterns emerge from the category breakdown. \textbf{Forbidden} instructions are the easiest category (99.1\% average), as models reliably avoid prohibited words and punctuation. \textbf{Pattern} (morphological constraints) and \textbf{Structure} are the hardest, averaging 84.8\% and 87.9\% respectively. Claude models show a striking weakness in \textbf{Structure} (37.5--59.4\%), particularly on acrostic constraints, while excelling in \textbf{Punctuation} and \textbf{Format}. This suggests that Claude models prioritize well-formatted responses but struggle with structural planning. The \textbf{Punctuation} category reveals an interesting split: GPT-5 family models score only 80--90\% (primarily failing on \texttt{include\_quote}), while most other models achieve 100\% on the same instructions. Portuguese morphological patterns remain challenging across all model families, with \texttt{terminacao\_inho\_inha\_min} (words ending in -inho/-inha) being particularly difficult for non-GPT models.

\subsection{Hardest Instructions}

Table~\ref{tab:hardest_instructions} identifies instruction types with lowest average accuracy across all models.

\begin{table}[ht]
\centering
\caption{Top 10 hardest instructions (average accuracy across all models).}
\label{tab:hardest_instructions}
\begin{tabular}{@{}lc@{}}
\toprule
\textbf{Instruction} & \textbf{Avg. Accuracy (\%)} \\
\midrule
count:exact\_word\_count & 21.1 \\
words:max\_word\_repeat & 31.6 \\
count:character\_count\_range & 47.4 \\
structure:acrostic & 66.3 \\
pattern:terminacao\_inho\_inha\_min & 67.8 \\
punctuation:include\_quote & 73.7 \\
count:word\_count\_range & 76.2 \\
words:word\_frequency & 78.9 \\
pattern:terminacao\_ando\_endo\_indo\_limit & 79.7 \\
count:exact\_line\_count & 82.1 \\
\bottomrule
\end{tabular}
\end{table}

The hardest instruction (\texttt{exact\_word\_count} at 21.1\%) requires generating responses with a precise word count, leaving no tolerance. \texttt{max\_word\_repeat} constraints (31.6\%) limit how many times any word can appear, requiring vocabulary diversity monitoring. Portuguese-specific patterns like \texttt{terminacao\_inho\_inha\_min} (67.8\%) show that morphological constraints remain challenging even for models with otherwise strong performance.

\subsection{Qualitative Error Analysis}

To understand failure modes beyond aggregate statistics, we examine concrete examples across instruction categories. Appendix~\ref{app:error_analysis} presents four representative failure cases with complete prompts and model responses, organized by error type: exact count failures (models lack reliable token-counting mechanisms during decoding), acrostic planning errors (markdown headers break structural patterns), morphological limit violations (gerund forms are natural in literary analysis, making limits particularly challenging), and formatting artifacts (models trained for well-formatted output insert markdown headers that violate start-word constraints). A recurring pattern emerges: models prioritize structural conventions (headers, formatting) over explicit instructions, suggesting that instruction-tuning habits can override specific user requirements. 
\section{Discussion}
\label{sec:discussion}

Our evaluation reveals several noteworthy patterns. First, reasoning-enabled models consistently outperform their base counterparts: GPT-5.2 improves from 91.5\% to 98.5\% strict accuracy with reasoning enabled, and smaller Qwen models with thinking (8B at 81.5\%, 32B at 82.5\%) match or exceed the larger 235B base variant (79.5\%). 

Second, Portuguese-specialized models achieve competitive performance. Sabiazinho-4 (87.0\%) surpasses Claude-Sonnet-4.5 (86.0\%) despite being a smaller, domain-specialized model, and outperforms Claude-Haiku-4.5 (73.5\%) at 8.6$\times$ lower cost (Appendix~\ref{app:cost}).

Third, the category-level analysis reveals that constraint difficulty is not uniform: forbidden instructions are near-trivially satisfied (99.1\% average), while morphological pattern constraints (84.8\%) and structural constraints (87.9\%) remain substantially harder. The punctuation category shows an unexpected model-specific split, with GPT-5 family models consistently underperforming on \texttt{include\_quote} despite excelling elsewhere.

\subsection{Limitations}
\label{sec:limitations}

\paragraph{Coherence Evaluation.} We use GPT-4o-mini as a coherence judge without human validation of the resulting scores. The high coherence across all models (91--99\%) suggests the judge may lack sufficient discriminative power, potentially failing to penalize subtle quality differences. A human annotation study correlating judge scores with expert assessments would strengthen this component of the evaluation.

\paragraph{Surface-Level Morphological Verification.} Because morphological pattern instructions are verified through suffix matching rather than semantic analysis, any word ending in the target suffix is accepted regardless of whether it constitutes a genuine instance of the grammatical category. For example, a word ending in \textit{-inho} is counted even if it is not a true diminutive, and a model-invented word matching the pattern would also pass verification. This is an inherent trade-off of the automatic verification principle: we gain full reproducibility and scalability at the cost of not distinguishing semantically valid uses from superficial pattern matches. In practice, the manual audit (Appendix~\ref{app:verification}) found no false positives attributable to this issue, suggesting that models generally produce real words when satisfying these constraints.

\section{Conclusion}
\label{sec:conclusion}

We presented CAPITU, the first instruction-following benchmark specifically designed for Brazilian Portuguese. By combining 59 deterministically verifiable instruction types with cultural grounding in eight canonical works of Brazilian literature, CAPITU enables comprehensive evaluation of LLM capabilities in Portuguese.

Our evaluation of 18 models reveals that frontier reasoning models can achieve strong instruction-following in Portuguese, with GPT-5.2 (reasoning-effort high) reaching 98.5\% strict accuracy. However, significant variation exists across model families, particularly in multi-turn settings where conversation accuracy ranges from 60\% to 96\%. Portuguese-specialized models offer compelling cost-performance trade-offs (Appendix~\ref{app:cost}): Sabiazinho-4 achieves 87.0\% accuracy at \$0.13, outperforming Claude-Haiku-4.5 at 8.6$\times$ lower cost. Key challenges remain in Portuguese morphological constraints (diminutives, gerunds), precise counting, structural planning, and constraint persistence across conversation turns.

We release the complete benchmark, evaluation code, and generation tools to support research on Portuguese NLP and instruction-following evaluation. Future work includes expanding the literary corpus to other Lusophone cultures, increasing the proportion of Portuguese-specific instructions, incorporating human validation of both verification functions and coherence scores, and investigating techniques to improve multi-turn constraint persistence.

\bibliographystyle{plainnat}
\bibliography{references}
\clearpage

\begin{appendices}
\appendix
\makeatletter
\renewcommand{\@seccntformat}[1]{Appendix \csname the#1\endcsname\quad}
\makeatother
\section{Complete Instruction List}
\label{app:instructions}

\begin{longtable}[c]{@{}lp{8cm}@{}}
\caption{Complete instruction types in CAPITU (59 total).}
\label{tab:full_instructions} \\
\toprule
\textbf{Instruction ID} & \textbf{Description} \\
\midrule
\endfirsthead

\multicolumn{2}{c}%
{\tablename\ \thetable\ -- \textit{Continued from previous page}} \\
\toprule
\textbf{Instruction ID} & \textbf{Description} \\
\midrule
\endhead

\midrule
\multicolumn{2}{r}{\textit{Continued on next page}} \\
\endfoot

\bottomrule
\endlastfoot

\multicolumn{2}{@{}l}{\textit{Count (17)}} \\
count:word\_count\_range & Response must have between X and Y words \\
count:exact\_word\_count & Response must have exactly N words \\
count:min\_word\_count & Response must have at least N words \\
count:max\_word\_count & Response must have at most N words \\
count:unique\_word\_count & Use at least N unique words \\
count:character\_count\_range & Response must have between X and Y characters \\
count:exact\_sentence\_count & Response must have exactly N sentences \\
count:min\_sentence\_count & Response must have at least N sentences \\
count:sentence\_count\_range & Response must have between X and Y sentences \\
count:exact\_paragraph\_count & Response must have exactly N paragraphs \\
count:min\_paragraph\_count & Response must have at least N paragraphs \\
count:max\_sentence\_length & Each sentence must have at most N words \\
count:min\_sentence\_length & Each sentence must have at least N words \\
count:exact\_line\_count & Response must have exactly N lines \\
count:exact\_number\_count & Use exactly N numbers \\
count:min\_number\_count & Use at least N numbers \\
count:include\_specific\_number & Include specific number N \\
\midrule
\multicolumn{2}{@{}l}{\textit{Words (11)}} \\
words:include\_word & Include specific word \\
words:include\_words & Include all words from list \\
words:word\_frequency & Use word exactly N times \\
words:min\_word\_frequency & Use word at least N times \\
words:use\_first\_person & Use first person pronouns \\
words:use\_third\_person & Use third person pronouns \\
words:conjunction\_count & Use at least N conjunctions \\
words:connective & Use at least N connectives \\
words:temporal\_marker & Use at least N temporal markers \\
words:contrast\_marker & Include at least one contrast marker \\
words:max\_word\_repeat & No word repeats more than N times \\
\midrule
\multicolumn{2}{@{}l}{\textit{Forbidden (7)}} \\
forbidden:word & Do NOT use specific word \\
forbidden:words\_list & Do NOT use any words from list \\
forbidden:no\_numbers & Do NOT use numerals \\
forbidden:no\_first\_person & Do NOT use first-person pronouns \\
forbidden:no\_second\_person & Do NOT use second-person pronouns \\
forbidden:no\_questions & Do NOT use question marks \\
forbidden:no\_exclamations & Do NOT use exclamation marks \\
\midrule
\multicolumn{2}{@{}l}{\textit{Punctuation (5)}} \\
punctuation:only\_declarative & Only declarative sentences (no ? or !) \\
punctuation:include\_question & Include at least one question \\
punctuation:include\_quote & Include quoted text ("...") \\
punctuation:use\_semicolon & Use at least N semicolons \\
punctuation:use\_colon & Use at least N colons \\
\midrule
\multicolumn{2}{@{}l}{\textit{Structure (6)}} \\
structure:start\_with\_word & Start with specific word \\
structure:end\_with\_word & End with specific word \\
structure:start\_end\_same\_word & Start and end with same word \\
structure:no\_repeat\_sentence\_start & No consecutive sentences start same \\
structure:each\_line\_starts\_with & Each line starts with character \\
structure:acrostic & First letters form specified word \\
\midrule
\multicolumn{2}{@{}l}{\textit{Format (5)}} \\
format:bullet\_list & Use bullet list format \\
format:numbered\_list & Use numbered list format \\
format:all\_caps & Write entirely in uppercase \\
format:all\_lowercase & Write entirely in lowercase \\
format:title\_case\_start & Capitalize first letter of each sentence \\
\midrule
\multicolumn{2}{@{}l}{\textit{Pattern (8)}} \\
pattern:terminacao\_ando\_endo\_indo\_limit & Max N words ending in -ando/-endo/-indo \\
pattern:terminacao\_ando\_endo\_indo\_min & Min N words ending in -ando/-endo/-indo \\
pattern:terminacao\_inho\_inha\_min & Min N words ending in -inho/-inha \\
pattern:terminacao\_inho\_inha\_proibido & No words ending in -inho/-inha \\
pattern:terminacao\_ao\_oes\_min & Min N words ending in -ão/-ões \\
pattern:terminacao\_mente\_limit & Max N words ending in -mente \\
pattern:terminacao\_mente\_min & Min N words ending in -mente \\
pattern:terminacao\_mente\_proibido & No words ending in -mente \\
\end{longtable}
\section{Example Prompts}
\label{app:examples}

\subsection{Single-Turn Example (Medium Difficulty)}

\begin{tcolorbox}[
    colback=gray!5,
    colframe=gray!75!black,
    title=Prompt,
    fonttitle=\bfseries
]
Escreva uma análise crítica de Dom Casmurro, de Machado de Assis, explorando como a obra aborda identidade e pertencimento. Siga estritamente: a resposta deve conter entre 120 e 150 palavras; e use no máximo 5 palavras terminadas em -mente.
\end{tcolorbox}

\vspace{0.3cm}
\noindent\textbf{Instructions:}
\begin{itemize}[leftmargin=1.5cm]
    \item \texttt{count:word\_count\_range} \textcolor{gray}{(min\_words=120, max\_words=150)}
    \item \texttt{pattern:terminacao\_mente\_limit} \textcolor{gray}{(max\_count=5)}
\end{itemize}

\subsection{Multi-Turn Example}

\begin{tcolorbox}[
    colback=blue!5,
    colframe=blue!75!black,
    title=Turn 1,
    fonttitle=\bfseries
]
Apresente brevemente a obra `Vidas Secas' de Graciliano Ramos. Siga estritamente: a resposta deve ter pelo menos 80 palavras.
\end{tcolorbox}

\vspace{0.3cm}
\begin{tcolorbox}[
    colback=blue!5,
    colframe=blue!75!black,
    title=Turn 2,
    fonttitle=\bfseries
]
Agora aprofunde a análise, focando em desigualdade social. Siga estritamente: a resposta deve ter pelo menos 80 palavras; e use pelo menos 2 conectivos (portanto, assim, além disso, etc.).
\end{tcolorbox}

\vspace{0.3cm}
\begin{tcolorbox}[
    colback=blue!5,
    colframe=blue!75!black,
    title=Turn 3,
    fonttitle=\bfseries
]
Para concluir, compare esta obra com outra do mesmo período literário. Siga estritamente: a resposta deve ter pelo menos 80 palavras; use pelo menos 2 conectivos; e NÃO use pronomes de primeira pessoa.
\end{tcolorbox}

\vspace{0.3cm}
\noindent\textbf{Cumulative Instructions:}
\begin{itemize}[leftmargin=1.5cm]
    \item[\textbf{T1:}] \texttt{count:min\_word\_count} \textcolor{gray}{(min\_words=80)}
    \item[\textbf{T2:}] + \texttt{words:connective} \textcolor{gray}{(min\_count=2)}
    \item[\textbf{T3:}] + \texttt{forbidden:no\_first\_person}
\end{itemize}

\section{Verification Implementation Details}
\label{app:verification_impl}

This appendix provides implementation details for the four verification categories described in Section~\ref{sec:methodology}.

\paragraph{Count-Based Verification}
Count instructions employ string operations with consistent tokenization rules. Word counting splits on whitespace after punctuation removal; sentence counting splits on terminal punctuation; paragraph counting splits on double newlines; character counting includes whitespace. 

\paragraph{Pattern-Based Verification}
Portuguese morphological constraints use regular expressions targeting specific word terminations. Gerund forms are detected via the pattern -ando/-endo/-indo; diminutives via -inho/-inha/-zinho/-zinha; and adverbs via -mente. All patterns operate on lowercased text with word boundary anchors to prevent partial matches within longer words.

\paragraph{Lexicon-Based Verification}
Constraints involving grammatical categories match against closed, manually curated lexicons. First-person prohibition checks against 16 Portuguese pronouns (\textit{eu}, \textit{me}, \textit{mim}, \textit{comigo}, \textit{meu}, \textit{minha}, \textit{nós}, \textit{nosso}, etc.). Connective requirements verify presence of discourse markers organized by function: logical (\textit{portanto}, \textit{assim}), adversative (\textit{contudo}, \textit{porém}), and additive (\textit{além disso}, \textit{ademais}). Temporal markers include sequential indicators such as \textit{primeiro}, \textit{depois}, \textit{por fim}, and \textit{finalmente}. Each lexicon entry is matched with word boundaries to avoid false positives from substrings.

\paragraph{Structural Verification}
Structural constraints verify text organization and formatting. Acrostic verification extracts the initial character of each sentence and compares the resulting string against the required word. Start-with-word constraints check the first token of the response. Quote inclusion detects both straight and typographic quotation marks. List format verification identifies bullet or numbering markers at line boundaries.

\section{Coherence Evaluation Prompt}
\begin{figure}[ht]
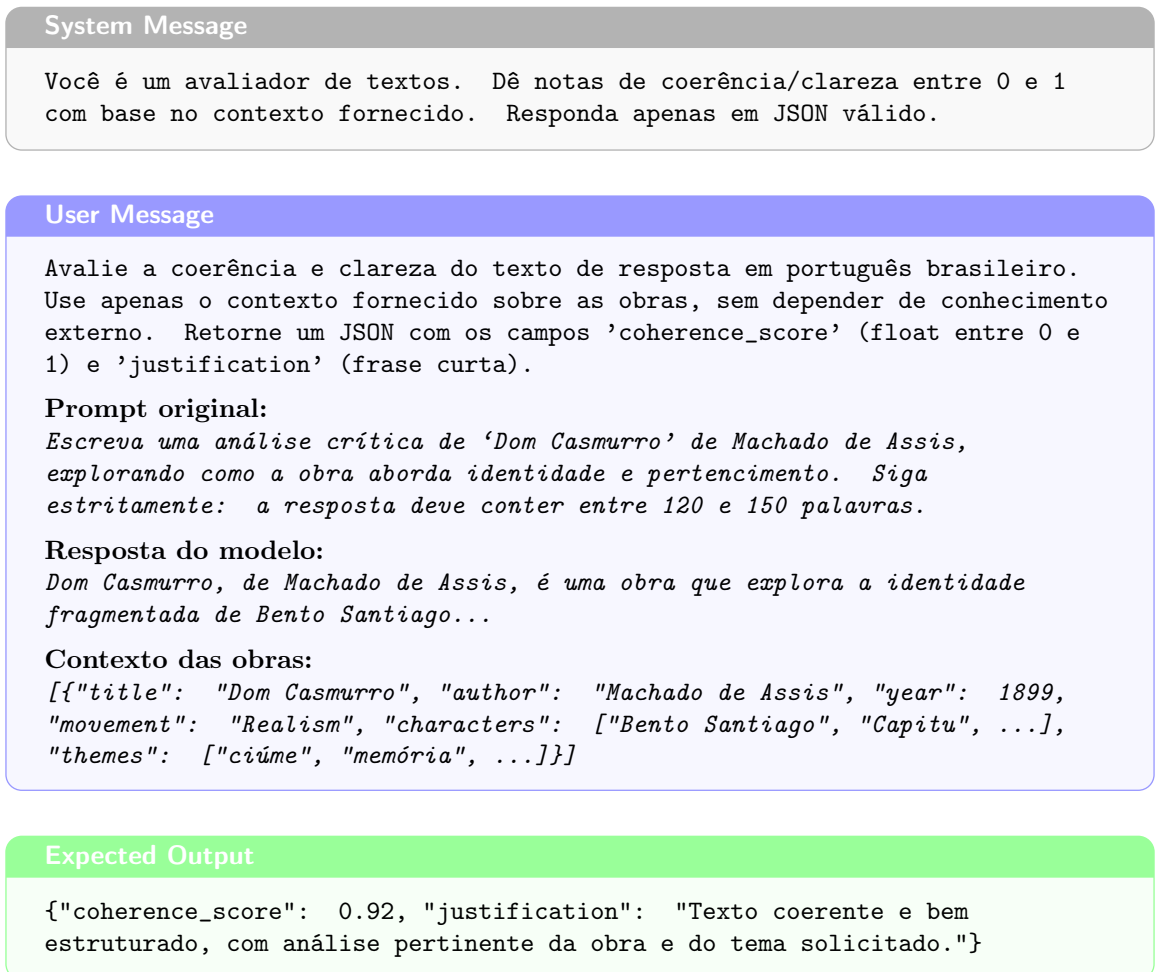

\centering

\begin{tcolorbox}[
    colback=gray!5,
    colframe=gray!60,
    width=0.95\textwidth,
    arc=2mm,
    boxrule=0.5pt,
    title={\small\textbf{System Message}},
    fonttitle=\sffamily
]
\small\ttfamily
Voc\^e \'e um avaliador de textos. D\^e notas de coer\^encia/clareza entre 0 e 1 com base no contexto fornecido. Responda apenas em JSON v\'alido.
\end{tcolorbox}

\vspace{0.3em}

\begin{tcolorbox}[
    colback=blue!3,
    colframe=blue!40,
    width=0.95\textwidth,
    arc=2mm,
    boxrule=0.5pt,
    title={\small\textbf{User Message}},
    fonttitle=\sffamily
]
\small\ttfamily
Avalie a coer\^encia e clareza do texto de resposta em portugu\^es brasileiro. Use apenas o contexto fornecido sobre as obras, sem depender de conhecimento externo. Retorne um JSON com os campos 'coherence\_score' (float entre 0 e 1) e 'justification' (frase curta).

\vspace{0.5em}
\rmfamily\small\textbf{Prompt original:}\\
\ttfamily\small\textit{Escreva uma an\'alise cr\'itica de `Dom Casmurro' de Machado de Assis, explorando como a obra aborda identidade e pertencimento. Siga estritamente: a resposta deve conter entre 120 e 150 palavras.}

\vspace{0.5em}
\rmfamily\small\textbf{Resposta do modelo:}\\
\ttfamily\small\textit{Dom Casmurro, de Machado de Assis, \'e uma obra que explora a identidade fragmentada de Bento Santiago...}

\vspace{0.5em}
\rmfamily\small\textbf{Contexto das obras:}\\
\ttfamily\small\textit{[\{"title": "Dom Casmurro", "author": "Machado de Assis", "year": 1899, "movement": "Realism", "characters": ["Bento Santiago", "Capitu", ...], "themes": ["ci\'ume", "mem\'oria", ...]\}]}
\end{tcolorbox}

\vspace{0.3em}

\begin{tcolorbox}[
    colback=green!3,
    colframe=green!40,
    width=0.95\textwidth,
    arc=2mm,
    boxrule=0.5pt,
    title={\small\textbf{Expected Output}},
    fonttitle=\sffamily
]
\small\ttfamily
\{"coherence\_score": 0.92, "justification": "Texto coerente e bem estruturado, com an\'alise pertinente da obra e do tema solicitado."\}
\end{tcolorbox}

\caption{Complete coherence evaluation prompt structure. The system message defines the judge's role; the user message provides the original prompt, the model's response, and literary metadata as context; the judge returns a JSON score with justification.}
\label{fig:coherence_prompt}
\end{figure}

Figure~\ref{fig:coherence_prompt} illustrates the complete prompt structure with a concrete example.

\section{Instruction Combinations by Difficulty}
\label{app:combinations}

This appendix lists the instruction combinations used for each difficulty level. All combinations are manually curated to ensure logical consistency.

\paragraph{Easy (14 single instructions).}
Each easy question uses one instruction randomly selected from: \texttt{min\_word\_count}, \texttt{max\_word\_count}, \texttt{min\_sentence\_count}, \texttt{min\_paragraph\_count}, \texttt{no\_questions}, \texttt{no\_exclamations}, \texttt{no\_numbers}, \texttt{title\_case\_start}, \texttt{bullet\_list}, \texttt{start\_with\_word}, \texttt{include\_word}, \texttt{connective}, \texttt{terminacao\_ando\_endo\_indo\_limit}, \texttt{terminacao\_mente\_limit}.

\paragraph{Medium (25 pairs).}
Each medium question uses a curated pair combining constraints from different categories. Representative pairs include:

\begin{itemize}[nosep,leftmargin=1.5cm]
\item \textit{Count + Forbidden}: \texttt{word\_count\_range} + \texttt{no\_questions}; \texttt{min\_word\_count} + \texttt{forbidden:word}
\item \textit{Count + Pattern}: \texttt{word\_count\_range} + \texttt{terminacao\_ando\_endo\_indo\_limit}; \texttt{min\_word\_count} + \texttt{terminacao\_mente\_limit}
\item \textit{Structure + Count}: \texttt{start\_with\_word} + \texttt{word\_count\_range}; \texttt{end\_with\_word} + \texttt{min\_sentence\_count}
\item \textit{Punctuation + Count}: \texttt{include\_quote} + \texttt{min\_sentence\_count}; \texttt{include\_question} + \texttt{word\_count\_range}
\item \textit{Pattern + Words}: \texttt{terminacao\_ando\_endo\_indo\_min} + \texttt{connective}; \texttt{terminacao\_mente\_min} + \texttt{include\_word}
\end{itemize}

\paragraph{Hard (23 combinations of 3--4 instructions).}
Hard questions combine three or four constraints that create competing demands. Examples include:

\begin{itemize}[nosep,leftmargin=1.5cm]
\item \textit{Triplets}: \texttt{exact\_word\_count} + \texttt{no\_questions} + \texttt{no\_exclamations}; \texttt{exact\_paragraph\_count} + \texttt{start\_with\_word} + \texttt{forbidden:word}; \texttt{terminacao\_inho\_inha\_min} + \texttt{terminacao\_mente\_limit} + \texttt{word\_count\_range}
\item \textit{Quadruplets}: \texttt{word\_count\_range} + \texttt{no\_questions} + \texttt{no\_first\_person} + \texttt{terminacao\_mente\_limit}; \texttt{acrostic} + \texttt{exact\_line\_count} + \texttt{no\_questions} + \texttt{terminacao\_mente\_proibido}; \texttt{exact\_word\_count} + \texttt{max\_word\_repeat} + \texttt{no\_first\_person} + \texttt{contrast\_marker}
\end{itemize}

The complete list of all 62 combinations (14 easy + 25 medium + 23 hard) is available in the source code repository.

\section{Cost-Performance Analysis}
\label{app:cost}

Table~\ref{tab:cost_performance} presents the total API cost for evaluating each model on both single-turn (200 prompts) and multi-turn (100 conversations) benchmarks, alongside performance metrics. Costs are computed from recorded token counts (prompt and completion tokens, including reasoning tokens where applicable) multiplied by each provider's published per-token rates at the time of evaluation.

\small
\setlength{\tabcolsep}{4pt}
\renewcommand{\arraystretch}{1.25}
\begin{xltabular}{\textwidth}{@{}Xrrrccr@{}}
\caption{Cost-performance analysis. \textbf{In} and \textbf{Out} = per-million token rates (USD) for input and output respectively. \textbf{Cost} = total API cost for the full benchmark (200 single-turn + 100 multi-turn). \textbf{Strict} = single-turn strict accuracy; \textbf{Conv.} = multi-turn conversation accuracy; \textbf{Perf/\$} = strict accuracy divided by cost. Models sorted by cost descending.}
\label{tab:cost_performance} \\
\toprule
\textbf{Model} & \textbf{In (\$/1M)} & \textbf{Out (\$/1M)} & \textbf{Cost (USD)} & \textbf{Strict (\%)} & \textbf{Conv. (\%)} & \textbf{Perf/\$} \\
\midrule
\multicolumn{7}{@{}l}{\textit{High Cost ($>$\$1)}} \\[2pt]
Gemini-3-Pro-Preview$^\dagger$ & 2.00 & 12.00 & \$10.80 & 92.5 & 95.0 & 8.6 \\
GPT-5$^\dagger$ & 1.25 & 10.00 & \$8.76 & 98.0 & 90.0 & 11.2 \\
GPT-5.2$^\dagger$ & 1.75 & 14.00 & \$5.04 & 98.5 & 83.0 & 19.5 \\
Qwen3-235b-a22b$^\dagger$ & 0.40 & 4.00 & \$3.72 & 84.0 & 64.0 & 22.6 \\
Claude-Sonnet-4.5 & 3.00 & 15.00 & \$3.69 & 86.0 & 87.0 & 23.3 \\
GPT-5.2 & 1.75 & 14.00 & \$2.51 & 91.5 & 73.0 & 36.5 \\
GPT-5-mini$^\dagger$ & 0.25 & 2.00 & \$1.36 & 98.0 & 96.0 & 72.1 \\
Claude-Haiku-4.5 & 1.00 & 5.00 & \$1.12 & 73.5 & 72.0 & 65.6 \\
\midrule
\multicolumn{7}{@{}l}{\textit{Medium Cost (\$0.30--\$1)}} \\[2pt]
Sabiá-4 & 0.96 & 3.85 & \$0.64 & 84.5 & 78.0 & 132.0 \\
Gemini-3-Flash-Preview & 0.50 & 3.00 & \$0.48 & 89.0 & 95.0 & 185.4 \\
Qwen3-32B$^\dagger$ & 0.29 & 1.15 & \$0.46 & 82.5 & 60.0 & 179.3 \\
Sabiá-3 & 0.96 & 1.92 & \$0.45 & 77.0 & 64.0 & 171.1 \\
Sabiá-3.1 & 0.96 & 1.92 & \$0.42 & 74.0 & 64.0 & 176.2 \\
Qwen3-14B$^\dagger$ & 0.29 & 1.15 & \$0.39 & 81.0 & 74.0 & 207.7 \\
\midrule
\multicolumn{7}{@{}l}{\textit{Low Cost ($<$\$0.30)}} \\[2pt]
Qwen3-8B$^\dagger$ & 0.18 & 0.70 & \$0.25 & 81.5 & 72.0 & 326.0 \\
Qwen3-235b-a22b & 0.29 & 1.15 & \$0.22 & 79.5 & 60.0 & 361.4 \\
Sabiazinho-4 & 0.19 & 0.77 & \$0.13 & 87.0 & 87.0 & 669.2 \\
Sabiazinho-3 & 0.19 & 0.77 & \$0.13 & 80.0 & 68.0 & 615.4 \\
\bottomrule
\multicolumn{7}{@{}l}{\footnotesize $^\dagger$ Model with reasoning/thinking enabled; GPT-5.2$^\dagger$ uses \texttt{reasoning\_effort=high}.} \\
\end{xltabular}

The cost analysis reveals an 83$\times$ cost range across models, from \$0.13 (Sabiazinho models) to \$10.80 (Gemini-3-Pro-Preview). Three cost-performance tiers emerge:

\paragraph{Premium tier ($>$\$1).} GPT-5-mini stands out as the best value in this tier: it matches GPT-5's strict accuracy (98.0\%) at 6.4$\times$ lower cost (\$1.36 vs \$8.76) and achieves the highest multi-turn conversation accuracy (96.0\%). GPT-5.2 with \texttt{reasoning\_effort=high} reaches the highest strict accuracy overall (98.5\%) at \$5.04---its higher cost relative to GPT-5.2 base (\$2.51) reflects the additional reasoning tokens generated, not a higher per-token rate.

\paragraph{Mid-range tier (\$0.30--\$1).} Gemini-3-Flash-Preview achieves the best balance in this tier, with 89.0\% strict accuracy and 95.0\% conversation accuracy at only \$0.48. Sabiá-4 (\$0.64) provides the highest accuracy (84.5\%) among Portuguese-specialized models in this range.

\paragraph{Budget tier ($<$\$0.30).} Sabiazinho-4 (\$0.13) delivers 87.0\% strict accuracy---outperforming Claude-Haiku-4.5 (73.5\% at \$1.12) at 8.6$\times$ lower cost, making it the most cost-efficient Portuguese-specialized model. Open-weight Qwen3-8B (\$0.25) and Qwen3-235B base (\$0.22) also offer strong performance per dollar, with Sabiazinho-4 achieving the highest Perf/\$ ratio (669.2) among all models.

\section{Performance Across Literary Works}
\label{app:literary}

Although literary context serves primarily as a grounding mechanism rather than an evaluation target, we observe modest variation in performance across works. Averaging strict accuracy across all 18 models, Macuna\'ima (89.8\%) and Iracema (88.4\%) yield the highest performance, while Vidas Secas (82.1\%) and O Corti\c{c}o (83.6\%) prove most challenging. However, the spread across literary works (7.7 pp) is substantially smaller than the spread across difficulty levels (40+ pp) or instruction categories (14+ pp), confirming that the literary context does not introduce significant evaluation bias. The relatively lower accuracy on Vidas Secas may reflect its minimalist prose style, which constrains the vocabulary available for satisfying morphological requirements.

\section{Qualitative Error Analysis}
\label{app:error_analysis}

This appendix presents representative failure cases with complete prompts and model responses. Each example illustrates a distinct error category identified in Section~\ref{sec:results}.

\subsection{Exact Count Failure}

The model produces a coherent and relevant response but fails to meet the exact word count, suggesting that models lack reliable token-counting mechanisms during decoding. Claude-Haiku-4.5 produced 148 words instead of the required 180.

\begin{tcolorbox}[
    colback=gray!5,
    colframe=gray!75!black,
    title=Prompt (Hard -- \textit{A Hora da Estrela}),
    fonttitle=\bfseries
]
\small
Escreva um coment\'ario sobre `A Hora da Estrela' para um grupo de leitura, destacando desigualdade social. Siga estritamente: a resposta deve ter \textbf{exatamente 180 palavras}; nenhuma palavra pode se repetir mais de 5 vezes; N\~AO use pronomes de primeira pessoa (eu, me, mim, n\'os, meu, nosso, etc.); e inclua pelo menos um marcador de contraste (por\'em, contudo, no entanto, por outro lado, etc.).
\end{tcolorbox}

\vspace{0.3em}

\begin{tcolorbox}[
    colback=red!3,
    colframe=red!50,
    title=Response -- Claude-Haiku-4.5 \hfill {\normalfont\small\textcolor{red!70!black}{\ding{55} 148/180 words}},
    fonttitle=\bfseries
]
\small
\# Reflex\~ao sobre `A Hora da Estrela'

A obra de Clarice Lispector apresenta Macab\'ea como s\'imbolo vivo das desigualdades que marcam a sociedade brasileira. A personagem, migrante nordestina, enfrenta pobreza extrema, explora\c{c}\~ao laboral e invisibilidade social em contexto urbano hostil.

Lispector constr\'oi narrativa que exp\~oe cruamente as disparidades econ\^omicas. Macab\'ea trabalha em condi\c{c}\~oes prec\'arias, ganha sal\'ario insuficiente e habita espa\c{c}o miser\'avel, refletindo realidade de milh\~oes de brasileiros marginalizados. Seus sonhos modestos---beleza, reconhecimento, amor---contrastam brutalmente com limita\c{c}\~oes impostas pela classe social.

No entanto, a obra transcende simples den\'uncia: questiona responsabilidade coletiva diante da desumaniza\c{c}\~ao. Narrador c\'umplice reconhece dist\^ancia entre sua condi\c{c}\~ao privilegiada e sofrimento alheio, problematizando leitura confort\'avel.

A linguagem fragmentada e provocativa intensifica desconforto proposital. Lispector recusa sentimentalismo f\'acil, for\c{c}ando leitores a confrontar pr\'oprias culpabilidades estruturais.

Livro permanece urgente: convida reflex\~ao profunda sobre mecanismos que perpetuam desigualdade, tornando quest\~ao social inescap\'avel para qualquer interpreta\c{c}\~ao genu\'ina de sua mensagem.
\end{tcolorbox}

\vspace{0.3em}
\noindent\textbf{Instructions:}
\begin{itemize}[nosep,leftmargin=1.5cm]
    \item \texttt{count:exact\_word\_count} \textcolor{red}{(target=180, actual=148)} \ding{55}
    \item \texttt{words:max\_word\_repeat} \textcolor{gray}{(N=5)} \ding{51}
    \item \texttt{forbidden:no\_first\_person} \ding{51}
    \item \texttt{words:contrast\_marker} \ding{51}
\end{itemize}

\subsection{Acrostic Planning Failure}

The model attempts to produce an acrostic but prepends a markdown header, breaking the structural pattern. This is a systematic failure: Claude-Haiku-4.5 exhibits this behavior on 5 different acrostic prompts.

\begin{tcolorbox}[
    colback=gray!5,
    colframe=gray!75!black,
    title=Prompt (Hard -- \textit{Macuna\'ima}),
    fonttitle=\bfseries
]
\small
Discuta a import\^ancia de `Macuna\'ima' para a literatura brasileira, considerando a religiosidade popular. Siga estritamente: escreva um \textbf{acr\'ostico onde a primeira letra de cada linha forma a palavra `C\'EU'}; a resposta deve ter exatamente 8 linhas; N\~AO fa\c{c}a perguntas (n\~ao use `?'); e N\~AO use palavras terminadas em -mente.
\end{tcolorbox}

\vspace{0.3em}

\begin{tcolorbox}[
    colback=red!3,
    colframe=red!50,
    title=Response -- Claude-Haiku-4.5 \hfill {\normalfont\small\textcolor{red!70!black}{\ding{55} acrostic broken by \texttt{\#} header}},
    fonttitle=\bfseries
]
\small
\colorbox{yellow!30}{\# Macuna\'ima e a Religiosidade Popular} \hfill \textcolor{red}{$\leftarrow$ first character is \texttt{\#}, not \texttt{C}}

\textbf{C}ria\c{c}\~ao revolucion\'aria de M\'ario de Andrade que reinventou a prosa brasileira\\
\textbf{\'E}ntidade m\'itica que incorpora santos, orix\'as e divindades ind\'igenas em sua narrativa\\
\textbf{U}niversalidade do sagrado popular transformou a literatura em espa\c{c}o de inclus\~ao cultural

\textbf{C}elebra\c{c}\~ao do sincretismo religioso reflete a alma do Brasil profundo e verdadeiro\\
\textbf{E}spa\c{c}o onde o pag\~ao, o cat\'olico e o africano convivem sem hierarquias r\'igidas\\
\textbf{U}mbigo do mundo liter\'ario que conecta todas as for\c{c}as espirituais da na\c{c}\~ao\\
\textbf{I}nova\c{c}\~ao formal que rompeu com padr\~oes europeus atrav\'es da oralidade sagrada
\end{tcolorbox}

\vspace{0.3em}
\noindent\textbf{Instructions:}
\begin{itemize}[nosep,leftmargin=1.5cm]
    \item \texttt{structure:acrostic} \textcolor{red}{(word=C\'EU, first char=\#)} \ding{55}
    \item \texttt{count:exact\_line\_count} \textcolor{gray}{(N=8)} \ding{51}
    \item \texttt{forbidden:no\_questions} \ding{51}
    \item \texttt{pattern:terminacao\_mente\_proibido} \ding{51}
\end{itemize}

\subsection{Morphological Limit Violation}

Gerund forms (-ando, -endo, -indo) are natural in Portuguese literary analysis, making explicit limits on their use particularly challenging. Qwen3-235b produced exactly double the allowed count.

\begin{tcolorbox}[
    colback=gray!5,
    colframe=gray!75!black,
    title=Prompt (Medium -- \textit{Iracema}),
    fonttitle=\bfseries
]
\small
Produza uma breve resenha cr\'itica de `Iracema', analisando o tratamento de desigualdade social. Siga estritamente: a resposta deve conter entre 80 e 114 palavras; e use \textbf{no m\'aximo 4 palavras terminadas em -ando, -endo ou -indo}.
\end{tcolorbox}

\vspace{0.3em}

\begin{tcolorbox}[
    colback=red!3,
    colframe=red!50,
    title=Response -- Qwen3-235b-a22b \hfill {\normalfont\small\textcolor{red!70!black}{\ding{55} 8/4 words in -ando/-endo/-indo}},
    fonttitle=\bfseries
]
\small
\textit{Iracema}, de Jos\'e de Alencar, idealiza o nascimento do Brasil por meio do amor entre um guerreiro ind\'igena e uma virgem tabajara, \underline{simbolizando} a mestiçagem. Apesar da beleza l\'irica, o romance oculta as desigualdades sociais reais, \underline{romanticizando} a domina\c{c}\~ao colonial. Iracema \'e representada como terra virgem, \underline{entregando}-se passivamente ao branco, \underline{refor\c{c}ando} hierarquias opressoras. O silenciamento da viol\^encia e do despojo \'e evidente, \underline{sustentando} um nacionalismo excluinte. A obra, \underline{construindo} uma identidade nacional, marginaliza vozes ind\'igenas e nega os conflitos hist\'oricos. Assim, ao inv\'es de \underline{confrontando} a desigualdade, naturaliza-a, \underline{ofuscando} realidades indesejadas na constru\c{c}\~ao do mito brasileiro.
\end{tcolorbox}

\vspace{0.3em}
\noindent\textbf{Instructions:}
\begin{itemize}[nosep,leftmargin=1.5cm]
    \item \texttt{count:word\_count\_range} \textcolor{gray}{(80--114)} \ding{51}
    \item \texttt{pattern:terminacao\_ando\_endo\_indo\_limit} \textcolor{red}{(max=4, actual=8)} \ding{55}
\end{itemize}

\subsection{Start-Word Formatting Artifact}

Models trained to produce well-formatted responses insert markdown headers before the actual text, violating start-with-word constraints even when the required word appears in the header itself. Claude-Haiku-4.5 exhibited this pattern on 13 different prompts.

\begin{tcolorbox}[
    colback=gray!5,
    colframe=gray!75!black,
    title=Prompt (Medium -- \textit{A Hora da Estrela}),
    fonttitle=\bfseries
]
\small
Apresente `A Hora da Estrela' de Clarice Lispector como mediador de um clube de leitura, focando em marginaliza\c{c}\~ao e exclus\~ao. Siga estritamente: \textbf{comece a resposta com a palavra `Observando'}; e a resposta deve conter entre 180 e 204 palavras.
\end{tcolorbox}

\vspace{0.3em}

\begin{tcolorbox}[
    colback=red!3,
    colframe=red!50,
    title=Response -- Claude-Haiku-4.5 \hfill {\normalfont\small\textcolor{red!70!black}{\ding{55} first token is \texttt{\#}, not ``Observando''}},
    fonttitle=\bfseries
]
\small
\colorbox{yellow!30}{\# Observando `A Hora da Estrela' como Espelho Social} \hfill \textcolor{red}{$\leftarrow$ header prepended}

Observando a obra de Clarice Lispector, `A Hora da Estrela' apresenta-se como ferramenta fundamental para discutirmos marginaliza\c{c}\~ao e exclus\~ao. A protagonista Macab\'ea encarna a invisibilidade social: migrante nordestina, pobre, analfabeta funcional, trabalhando como datil\'ografa em condi\c{c}\~oes prec\'arias no Rio de Janeiro.

O romance revela como sistemas sociais exp\~oem os vulner\'aveis. Macab\'ea existe \`a margem, ignorada at\'e pelo narrador que a apresenta como personagem desprez\'ivel. Sua exclus\~ao n\~ao \'e acidental: \'e estrutural, refletindo desigualdades de classe, g\^enero e origem geogr\'afica.

Para nosso clube de leitura, sugiro quest\~oes provocadoras: Como a sociedade invisibiliza pessoas como Macab\'ea? Qual \'e nossa responsabilidade frente ao sofrimento alheio? O texto nos implica moralmente.

A novela denuncia a desumaniza\c{c}\~ao urbana e a indifença coletiva. Lispector for\c{c}a-nos a confrontar nossos preconceitos, reconhecendo que toda exclus\~ao come\c{c}a com a nega\c{c}\~ao de humanidade.

`A Hora da Estrela' transcende fic\c{c}\~ao: \'e convite ao engajamento social e \`a empatia genu\'ina com os marginalizados.
\end{tcolorbox}

\vspace{0.3em}
\noindent\textbf{Instructions:}
\begin{itemize}[nosep,leftmargin=1.5cm]
    \item \texttt{structure:start\_with\_word} \textcolor{red}{(word=Observando, actual first token=\#)} \ding{55}
    \item \texttt{count:word\_count\_range} \textcolor{red}{(180--204, actual=160)} \ding{55}
\end{itemize}

\section{Verification Validity}
\label{app:verification}

To assess the reliability of our automatic verification functions, we conducted a manual audit of 100 sampled evaluation cases. The sample was drawn from six models spanning different performance tiers (GPT-5.2 with reasoning, Claude-Sonnet-4.5, Gemini-3-Flash-Preview, Qwen3-235b, Sabiazinho-4, and Sabiá-3.1), stratified by pass/fail verdict (47 auto-FAIL, 53 auto-PASS), instruction category, and difficulty level. A human annotator independently assessed whether each focused instruction was correctly satisfied, without access to the automatic verdict.

The overall agreement rate between human judgment and automatic verification was \textbf{92\%} (92/100 cases). All 8 disagreements were \textit{false negatives}: the automatic system judged FAIL where the human judged PASS. No false positives were found (0/53 auto-PASS cases were overturned), indicating that when the system reports an instruction as satisfied, it is reliable. The 8 false negatives cluster in three verification functions:

\begin{itemize}[nosep]
    \item \texttt{format:bullet\_list} (5 cases): Responses were correctly formatted as bullet lists, but the checker's internal minimum item count (randomly set during generation) exceeded the number of items produced. Since the prompt instructs the model to ``use bullet format'' without specifying a minimum count, the human annotator judged the format constraint as satisfied.
    \item \texttt{forbidden:no\_first\_person} (2 cases): The word ``nos'' appeared as a contraction of ``em + os'' (e.g., ``nos momentos''), not as a first-person pronoun. The disambiguation heuristic failed to recognize these instances as contractions.
    \item \texttt{words:connective} (1 case): The response contained the required number of connectives, but the lexicon-based matcher did not detect all valid connective forms.
\end{itemize}

These results have two implications. First, the absence of false positives (precision = 100\%) means that reported accuracy scores are \textit{lower bounds}: true model performance is at least as high as measured. Second, the identified false negatives suggest specific improvements to the verification functions---expanding the contraction disambiguation heuristic, aligning the bullet list checker's expectations with the prompt wording, and extending the connective lexicon. Based on the observed 8\% false negative rate, we estimate that correcting these issues would increase reported strict accuracy by approximately 1--2 percentage points for most models.
\end{appendices}

\end{document}